\documentclass[10pt,journal,compsoc]{IEEEtran}

\usepackage{ragged2e}
\ifCLASSOPTIONcompsoc

\usepackage[nocompress]{cite}
\else

\usepackage{cite}
\fi

\usepackage{url} 

\usepackage{times}
\usepackage{graphicx} 
\usepackage{subfigure} 

\usepackage{algorithm}
\usepackage{algorithmic}

\usepackage{hyperref}

\usepackage{helvet}
\usepackage{courier}

\usepackage{makecell}
\usepackage{graphicx}
\usepackage{subfigure}
\usepackage{color}
\usepackage{amsfonts}
\usepackage{amsmath}
\usepackage{amssymb}

\usepackage{multirow}
\usepackage{booktabs}

\def\mB{{\mathcal B}}

\def\mD{{\mathcal D}}

\DeclareMathAlphabet\mathbfcal{OMS}{cmsy}{b}{n}

\def\0{{\bf 0}}
\def\1{{\bf 1}}

\def\bb{{\bf b}}

\def\bb{{\bf b}}

\def\ie{\mbox{\textit{i.e.}}}
\def\eg{\mbox{\textit{e.g.}}}
\def\wrt{\mbox{\textit{w.r.t. }}}

\usepackage{ntheorem}

\newtheorem*{*thm}{Theorem}

\newtheorem*{*lemma}{Lemma}
\newtheorem{remark}{Remark}

\usepackage{enumitem}

\def\sss{\scriptscriptstyle}

\newcommand{\eat}[1]{}
\usepackage{xspace}
\newcommand{\sexyname}{PNAG\xspace}

\definecolor{cyfcolor}{RGB}{14, 98, 81}

\def\mytitle{Pareto-aware Neural Architecture Generation for Diverse Computational Budgets}

\begin{document}
	
\title{\mytitle}
	
\author{Yong Guo, Yaofo Chen, Yin Zheng, Qi Chen, Peilin Zhao, Jian Chen, Junzhou Huang, Mingkui Tan*
\IEEEcompsocitemizethanks{
\IEEEcompsocthanksitem{Yong Guo is with with the School of Software Engineering, South China University of Technology. Yong Guo is also with Max Planck Institute for Informatics. E-mail: guo.yong@mail.scut.edu.cn}
\IEEEcompsocthanksitem{Yaofo Chen is with the School of Software Engineering, South China University of Technology. 
E-mail: sechenyaofo@mail.scut.edu.cn}
\IEEEcompsocthanksitem{Yin Zheng is with the Weixin Group, Tencent, China. E-mail: yzheng3xg@gmail.com}
\IEEEcompsocthanksitem{Qi Chen is with the School of Computer Science, the University of Adelaide. E-mail: qi.chen04@adelaide.edu.au}
\IEEEcompsocthanksitem{Peilin Zhao and Junzhou Huang are with Tencent AI Lab, Tencent, China. E-mail: \{masonzhao, joehhuang\}@tencent.com}
\IEEEcompsocthanksitem{Mingkui Tan and Jian Chen are with the School of Software Engineering, South China University of Technology.
Mingkui Tan is also with the Key Laboratory of Big Data and Intelligent Robot (South China University of Technology), Ministry of Education.
Mingkui Tan is also with the Pazhou Laboratory, Guangzhou, China.
E-mail: \{mingkuitan, ellachen\}@scut.edu.cn}
	\IEEEcompsocthanksitem{
	$^*$ Corresponding author}
}
}

\markboth{Journal of \LaTeX\ Class Files, 2022}%
{Shell \MakeLowercase{\textit{et al.}}: \mytitle}
	
\IEEEtitleabstractindextext{%
\begin{abstract}
\justifying
Designing feasible and effective architectures under diverse computational budgets, incurred by different applications/devices, is essential for deploying deep models in real-world applications.
To achieve this goal, existing methods often perform an independent architecture search process for each target budget, which is very inefficient yet unnecessary.
More critically, these independent search processes cannot share their learned knowledge (\ie, the distribution of good architectures) with each other and thus often result in limited search results.
To address these issues, 
we propose a Pareto-aware Neural Architecture Generator (\sexyname) which only needs to be trained once and dynamically produces the Pareto optimal architecture for any given budget via inference.
To train our \sexyname, we learn the whole Pareto frontier by jointly finding multiple Pareto optimal architectures under diverse budgets. 
Such a joint search algorithm not only greatly reduces the overall search cost but also improves the search results.
Extensive experiments on three hardware platforms (\ie, mobile device, CPU, and GPU) show the superiority of our method over existing methods.
\end{abstract}
		
}

\maketitle

\IEEEdisplaynontitleabstractindextext

\IEEEpeerreviewmaketitle

\IEEEraisesectionheading{\section{Introduction}\label{sec:introduction}}
	
\IEEEPARstart{D}{eep} neural networks (DNNs)~\cite{lecun1989backpropagation} have been the workhorse of many challenging tasks, including image classification~\cite{krizhevsky2012imagenet,srivastava2015training,he2016deep,alexey2021vit,liu2021swin}, semantic segmentation~\cite{evan2017fully,chen2018encoder,xie2021segformer,wang2021hrnet} and object detection~\cite{joseph2018yolov3,zhao2019object,tian2019fcos,nicolas2020end}.
However, designing effective architectures often relies heavily on human expertise.
To alleviate this issue, neural architecture search (NAS) methods have been proposed to automatically design effective architectures~\cite{zoph2016neural}. Existing studies show that these automatically searched architectures often outperform the manually designed ones in many computer vision tasks~\cite{zoph2018learning,li2020block,tan2020efficientdet,dai2021fbnetv3,white2021powerful,chen2021neural,yan2021fp,guo2020breaking}.

However, the state-of-the-art deep networks often contain a large number of parameters and come with extremely high computational cost.
As a result, it is hard to deploy these models to real-world scenarios with limited computation resources.
Regarding this issue, we have to carefully design architectures to fulfill a specific computational budget (\eg, a feasible model should have a latency lower than 100ms on a specified mobile device).
More critically, we may have to consider different computational budgets in the real world.
For example, a company may simultaneously develop/maintain multiple applications and each of them has a specific budget of latency. 

In order to design feasible architectures, most methods~\cite{tan2019mnasnet,stamoulis2019single} only considers a single computational budget and incorporates architecture's computational cost into the objective function of NAS.
When we consider diverse budgets, they have to conduct an independent search process for each budget~\cite{tan2019mnasnet}, which is very inefficient yet unnecessary.
Unlike these methods, one can also exploit the population-based methods to simultaneously find multiple architectures and then select an appropriate one from them to fulfill a specific budget~\cite{lu2019nsga,lu2020nsganetv2}. However, due to the limited population size, these searched architectures do not necessarily satisfy the required budget. More critically, all these searched architectures are fixed after search and cannot be easily adapted for a slightly changed budget. Thus, how to design effective architectures under diverse computational budgets in an efficient and flexible way still remains an open question.

\begin{figure*}
    \centering
	\subfigure[An illustration of generating feasible architectures for diverse budgets using PNAG.]{
		\includegraphics[width = 1.1\columnwidth]{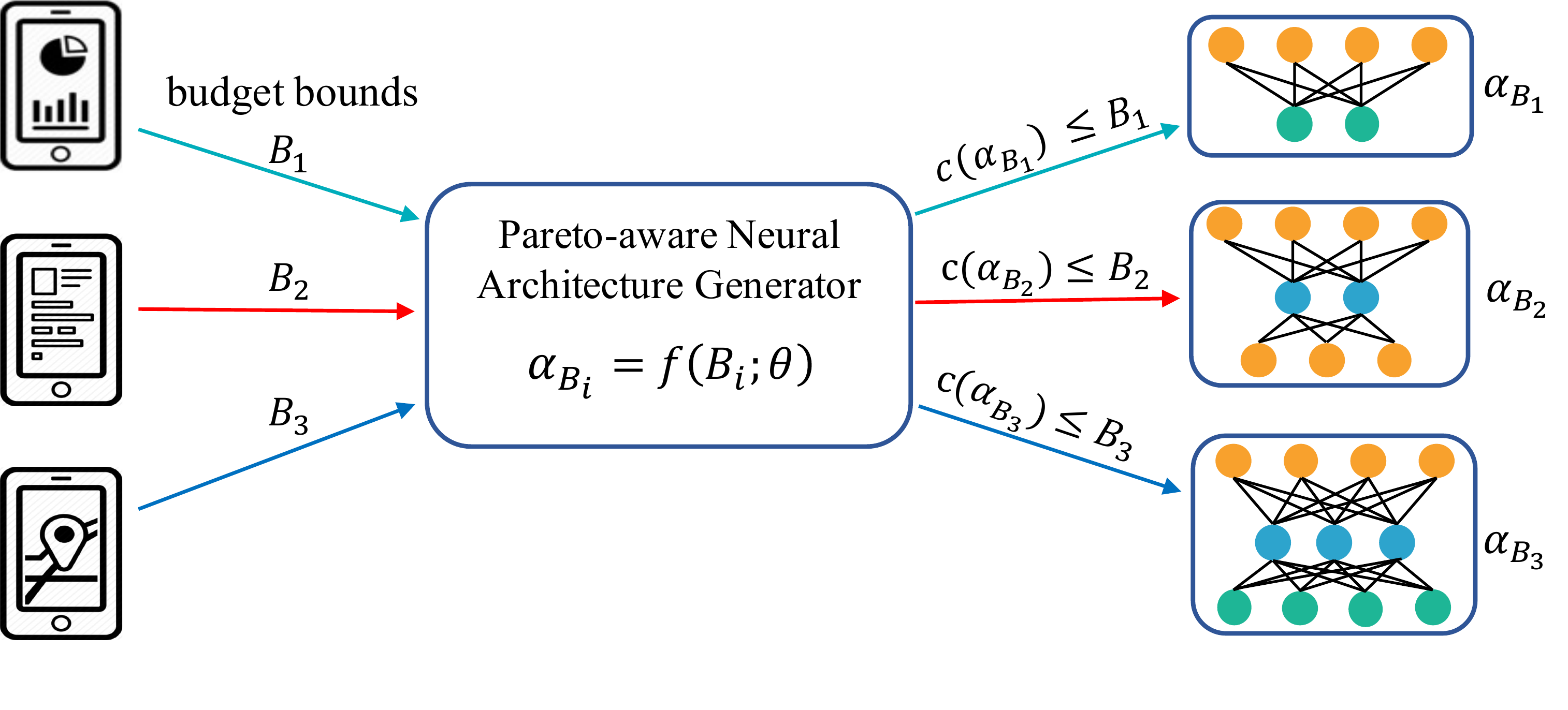}\label{fig:application_sub}
	}~
	\subfigure[Comparisons between PNAG and conventional NAS methods.]{
		\includegraphics[width = 0.82\columnwidth]{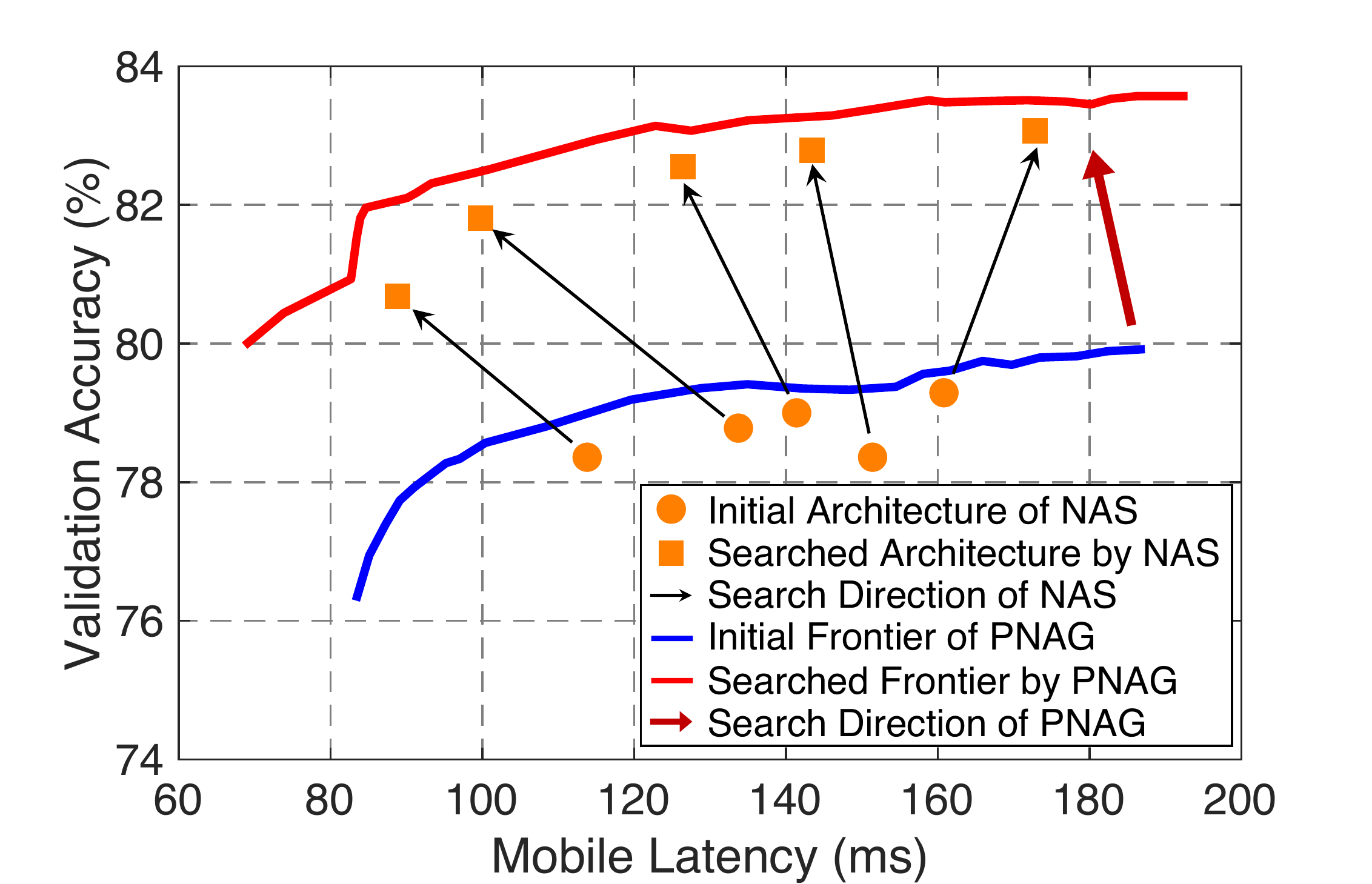}\label{fig:search_direction}
	}
    \caption{We show an illustration of how to apply PNAG to generate feasible architectures for diverse computational budgets and the comparisons between PNAG and conventional NAS methods. (a) \sexyname takes an arbitrary budget as input and flexibly generates architectures.
    (b) \sexyname learns the whole Pareto frontier rather than finding discrete architectures. Here, the accuracy is measured on the constructed validation set.
    }
    \label{fig:application}
\end{figure*}

In this paper, 
we propose a Pareto-aware Neural Architecture Generator (\sexyname) which only needs to be trained once and then dynamically produces Pareto optimal architectures for diverse budgets via \emph{inference} (as shown in Fig.~\ref{fig:application_sub}).
Note that the Pareto optimal architectures under different budgets should lie on a distribution, \ie, the Pareto frontier over model performance and computational cost~\cite{kim2005adaptive}.
We propose to jointly learn the whole Pareto frontier (\ie, improving the blue curve to the red curve in Fig.~\ref{fig:search_direction}) instead of finding a single Pareto optimal architecture.
During training, we randomly sample budgets from a predefined distribution and maximize the expected reward of the searched architectures to approximate the ground-truth Pareto frontier.
It is worth noting that learning the Pareto frontier is able to share the learned knowledge across different budgets and greatly improve the search results in practice (see results in Table~\ref{tab:pareto_learning}).
Furthermore, when evaluating architectures under diverse budgets, 
we design an architecture evaluator that learns a Pareto dominance rule to determine which architecture is a relatively better one in pairwise comparisons.
Unlike existing methods, we highlight that our PNAG designs architectures through a generation process instead of search, which is very efficient (see results in Table~\ref{tab:generation_cost}) and practically useful in real-world model deployment.


We summarize the contributions of our paper as follows.
\begin{itemize}
    \item 
    Instead of designing architectures for a single budget, we propose a Pareto-aware Neural Architecture Generator (\sexyname) which is only trained once and flexibly generates effective architectures for arbitrary budget via inference (see Fig.~\ref{fig:application_sub}). In this way, our architecture generation process becomes very efficient and practically useful in real-world applications.

    \item 
    To train our PNAG, we propose to explicitly learn the Pareto frontier by maximizing the expected reward of the searched architectures over diverse budgets. Interestingly, learning the Pareto frontier shares the learned knowledge across the search processes under diverse budgets and greatly improves the search results (see results in Table~\ref{tab:pareto_learning}).

    \item 
    Since an architecture should have different rewards/scores under different budgets, we propose an architecture evaluator to adaptively evaluate architectures for any given budget.
    To train the evaluator, we propose to learn a Pareto dominance rule which determines whether an architecture is better than the other in pairwise comparisons. 

    \item We measure the latencies on three hardware platforms and take them as the computational budgets to generate feasible architectures. Extensive experiments show that the architectures produced by \sexyname consistently outperform the architectures searched by existing methods across different budgets and platforms. 
\end{itemize}

\section{Related Work}

In this section, we provide a brief overview of existing work on neural architecture search, architecture design under resource constraints, as well as Pareto frontier learning.

\subsection{Neural Architecture Search (NAS)} 
Unlike manually designing architectures with expert knowledge, NAS seeks to automatically design more effective architectures~\cite{he2020milenas,li2020sgas,yang2021netadaptv2, zhang2021you,zheng2021migonas}.
Existing NAS methods can be roughly divided into three categories, namely, reinforcement-learning-based methods, evolutionary approaches, and gradient-based methods.
Specifically, reinforcement-learning-based methods~\cite{zoph2016neural, pham2018efficient, pasunuru2019continual, tian2020offrl,arash2020unas} learn a controller to produce architectures. Evolutionary approaches~\cite{real2017large, real2019regularized,lu2021neural,ming_zennas_iccv2021,chen2021autoformer,liu2021survey} search for promising architectures by gradually evolving a population. 
Gradient-based methods~\cite{liu2018darts, chen2019progressive, xu2020pcdarts, chu2021dartsminus,chen2021drnas,guo2022towards} relax the search space to be continuous and optimize architectures by gradient descent.
Besides designing effective search algorithms, many efforts have also been made to improve the accuracy of architecture evaluation~\cite{zhao2021few,yi2021renas, chu2021fairnas}.
Unlike these methods that find a single architecture, one can design different architectures by training an architecture generator. Specifically, RandWire~\cite{xie2019exploring} designs stochastic network generators to generate randomly wired architectures.
NAGO~\cite{ru2020neural} is the first work to learn an architecture generator and proposes a hierarchical and graph-based search space to reduce the optimization difficulty.
However, these generated architectures tend to perform very similarly (\ie, low diversity) in terms of both model performance and computational cost~\cite{xie2019exploring,ru2020neural}.
Thus, these architectures may not satisfy an arbitrary required budget.
In other words, they still have to learn a generator for a required budget to produce feasible architectures.

\begin{figure*}[t]
    \centering
    \includegraphics[width=0.90\textwidth]{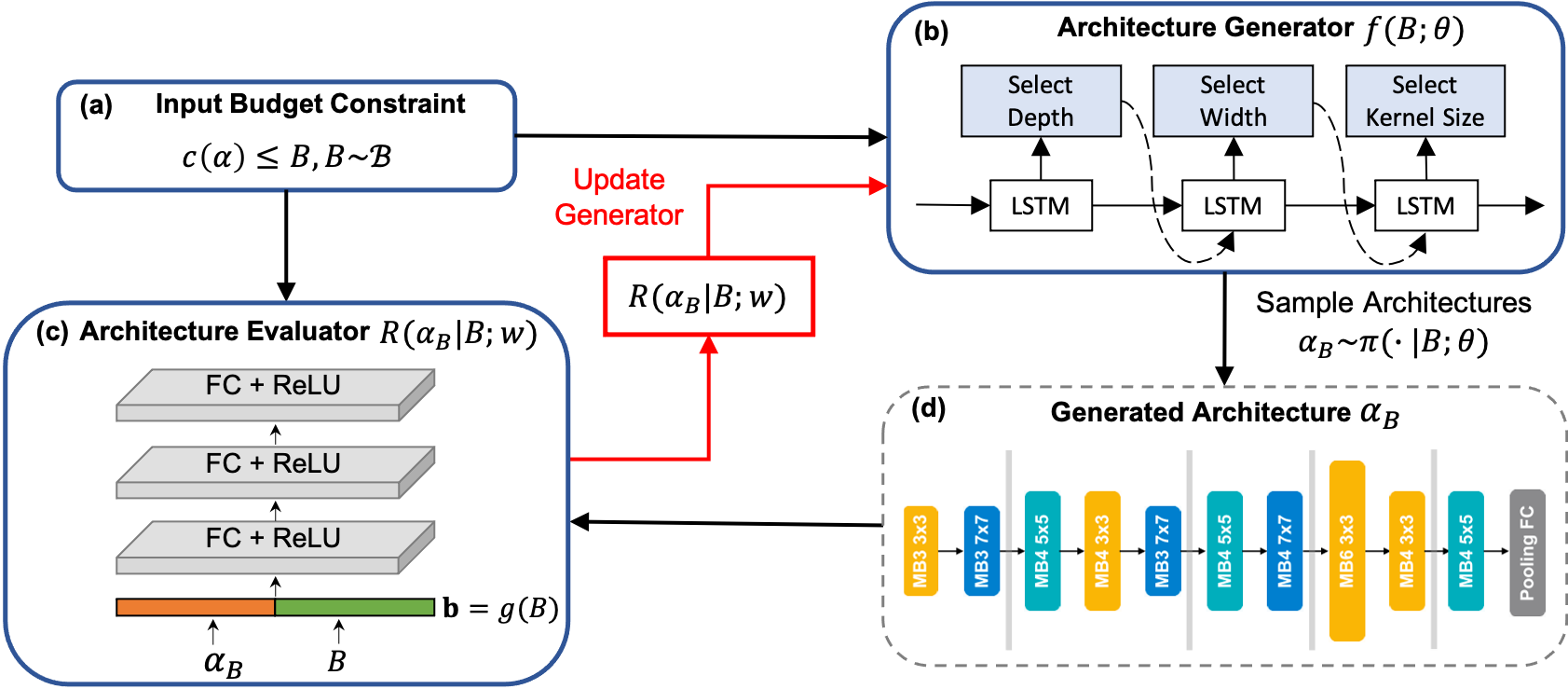}
    \caption{Overview of the proposed \sexyname. 
    Our \sexyname mainly consists of two modules: an architecture generator $f(\cdot;\theta)$ and an architecture evaluator $R(\cdot|\cdot;w)$. Specifically, we build the generator model based on an LSTM network, which takes a budget constraint $B$ as input and produces a promising architecture $\alpha_B$ that satisfies the budget constraint, \ie, $c(\alpha)$. To optimize the generator model, we design the evaluator using three fully connected (FC) layers to estimate the performance of the generated architectures $\alpha_B$. The orange and green boxes in (c) denote the embeddings of architecture $\alpha_{\sss B}$ and the budget w.r.t. $B$, respectively. }
    \label{fig:overview}
\end{figure*}

\subsection{Architecture Design under Resource Constraints} 
Many efforts have been made in designing architectures under a resource constraint~\cite{cai2019once, huang2020ponas, elsken2018efficient, Bender2020TuNAS, guo2020single,li2021hw}.
Specifically, PONAS~\cite{huang2020ponas} builds an accuracy table to find architectures satisfying a single budget constraint.
{TuNAS~\cite{Bender2020TuNAS} proposes a reward function to restrict the latency of the searched architecture, which omits additional hyper-parameter tuning.}
Related to our work, 
SGNAS~\cite{huang2021searching} proposes an architecture generator which generates architectures for specific budget constraints. Nevertheless, SGNAS optimizes a regression loss \wrt budget constraint and the resultant architecture does not necessarily have lower cost than the target budget, \ie, violating the budget. More critically, SGNAS considers a fixed hyper-parameter $\lambda$ to balance the regression loss and a classification loss. Due to the large diversity among architectures, their accuracy and computational cost may vary significantly across different budgets, also leading to suboptimal search results (See Table~\ref{tab:mobile_comp}).

\subsection{Pareto Frontier Learning}
Given multiple objectives, Pareto frontier learning aims to find a set of Pareto optimal solutions over them.
Most methods exploit evolutionary algorithms~\cite{deb2002fast,kim2004spea} to solve this problem.
Inspired by them, many efforts have been made to simultaneously find a set of Pareto optimal architectures over accuracy and computational cost~\cite{cheng2018searching,dong2018dpp}.
Recently, NSGANetV1~\cite{lu2020multi} presents an evolutionary approach to find a set of trade-off architectures over multiple objectives in a single run.
NSGANetV2~\cite{lu2020nsganetv2} further presents two surrogates (at the architecture and weights level) to produce task-specific models under multiple competing objectives.
Given a target budget, these methods may manually select an appropriate architecture from a set of searched architectures.
However, given limited population size, the selected architectures do not necessarily satisfy a required budget.
More critically, all the searched architectures are fixed after search and cannot be easily adapted for a slightly changed budget.
Thus, how to learn the Pareto frontier and use it to generate architectures for arbitrary budget in a flexible way still remains unexplored.

\section{Pareto-aware Architecture Generation}\label{sec:method}

In this paper, we focus on the architecture generation problem and intend to generate effective architectures for diverse computational budgets via \emph{inference} instead of search/training. 
{Note that the optimal architectures under different budgets lie on the Pareto frontier over model performance and computational cost~\cite{kim2005adaptive}. 
Thus, we develop a Pareto-aware Neural Architecture Generator (\sexyname) to explicitly learn the whole Pareto frontier.}
To locate the best architecture from the frontier for a given budget, we build our PNAG as a conditional model which takes the budget as input and directly produces a feasible architecture.
In Section~\ref{sec:generator}, we depict our architecture generator model and present a novel learning algorithm to learn the Pareto frontier. In Section~\ref{sec:reward}, we propose an architecture evaluator, as well as its training algorithm, to adaptively evaluate architectures under different budgets.
Algorithm~\ref{alg:training} shows the whole training process of \sexyname.

\subsection{Learning the Architecture Generator $f(B;\theta)$} \label{sec:generator}

We seek to build an architecture generator model {to dynamically and flexibly produce effective architectures for any given computational budget.}
Let $B$ be a budget (\eg, latency or MAdds) which can be considered as a random variable drawn from some distribution $\mB$, namely $B {\sim} \mB$.
Let $\Omega$ be an architecture search space. For any architecture $\alpha \in \Omega$, we use $c(\alpha)$ and ${\rm Acc}(\alpha)$ to measure the cost and validation accuracy of $\alpha$, respectively.

Since an architecture can be represented as a sequence of tokens (each token denotes a setting of a layer, \eg, width or kernel size)~\cite{zoph2016neural,pham2018efficient}, we cast the architecture generation problem as a sequential decision problem and build the architecture generator $f(B;\theta)$ using an LSTM network. 
As shown in Fig.~\ref{fig:overview}, the generator takes a budget $B$ as input and generates architectures $\alpha_{\sss B} {=} f(B;\theta)$ (satisfying the constraint $c(\alpha_{\sss B}) \leq B$) by sequentially predicting the token sequences, \ie, the depth, width, and kernel size of each layer.
Here, $\theta$ denotes the learnable parameters.
Note that the optimal architecture under a specific budget should lie on the Pareto frontier over model performance and computational cost.
To make the generator generalize to arbitrary budget, we seek to learn the Pareto frontier rather than finding discrete architectures. In the following, we first illustrate our training method in Section~\ref{sec:train_generator} and then discuss how to represent a budget with arbitrary value in Section~\ref{sec:rep_budget}.

\subsubsection{Training Method of $f(B;\theta)$}
\label{sec:train_generator}

To illustrate the training objective of our method, we first revisit the NAS problem with a single budget and then generalize it to the problem with diverse budgets.

\textbf{NAS under a single budget.}
Since it is non-trivial to directly find the optimal architecture~\cite{zoph2016neural}, by contrast, 
one can first learn a policy $\pi(\cdot; \theta)$ and then conduct sampling from it to find promising architectures, \ie, $\alpha \sim \pi(\cdot; \theta)$. Given a budget $B$, the optimization problem becomes
\begin{equation}\label{eq:obj-single-constraint}
    \begin{aligned}
         \max_{\theta} ~\mathbb{E}_{\alpha \sim \pi(\cdot; \theta)} ~\left[R \left( \alpha|B; w \right)\right], ~\text{s.t. } ~c(\alpha) \leq B.
     \end{aligned}
\end{equation}
Here, $\pi(\cdot;\theta)$ is the learned policy parameterized by $\theta$, and $R(\alpha|B; w)$ is the reward function parameterized by $w$ that measures the joint performance of both the accuracy and latency of $\alpha$. $\mathbb{E}_{\alpha \sim \pi(\cdot; \theta)} \left[  \cdot \right]$ is the expectation over the searched architectures. 

\begin{algorithm}[t]
\small
	\caption{Training method of \sexyname.}
	\label{alg:training}
	\begin{algorithmic}[1]\small
		\REQUIRE{
		Search space $\Omega$, latency distribution $\mB$, 
		learning rate $\eta$, training data set $\mD$, parameters $M$, $N$, and $K$.
		}
        \STATE Initialize model parameters $\theta$ for the generator and $w$ for the architecture evaluator. \\
        // \emph{Collect the architectures with accuracy and latency} \\
        \STATE Train a supernet $S$ on $\mD$. \\
        \STATE Randomly sample architectures $\left\{ \beta_i \right\}_{i=1}^{M}$ from $\Omega$. \\
        \STATE Construct tuples $\left\{( \beta_i, c(\beta_i), {\rm Acc}(\beta_i)) \right\}_{i=1}^{M}$ using $S$. \\
        // \emph{Learn the architecture evaluator} \\
        \WHILE{not convergent}
            \STATE Sample a set of latencies $\{B_k\}_{k=1}^{K}$ from $\mB$. \\
            \STATE Update the architecture evaluator by: \\
            \STATE ~~~~~~~~~$w \leftarrow w - \eta \nabla_w L(w)$. \\
        \ENDWHILE \\
        // \emph{Learn the architecture generator}  \\
        \WHILE{not convergent}
            \STATE Sample a set of latencies $\{B_k\}_{k=1}^{K}$ from $\mB$. \\
            \STATE Obtain $\{\alpha_{\sss {B_k}}^{\sss (i)}\}_{i=1}^{N}$ from $\pi(\cdot|B_k; \theta)$ for each $B_k$. \\
            \STATE Update the generator via policy gradient by: \\
            \STATE ~~~~~~~~~$\theta \leftarrow \theta + \eta \nabla_\theta J(\theta)$. \\
        \ENDWHILE
    \end{algorithmic}
\end{algorithm}

\textbf{NAS under diverse budgets.}
Problem~(\ref{eq:obj-single-constraint}) only focuses on one specific budget constraint. In fact, we seek to learn the Pareto frontier over the whole range of budgets (\eg, latency).
However, this problem is hard to solve since there may exist infinite Pareto optimal architectures with different computational cost. To address this, one can learn an approximated Pareto frontier by finding a set of uniformly distributed Pareto optimal points~\cite{grosan2008generating}. Here, we evenly sample $K$ budgets from the range of latency and maximize the expected reward over them.
Thus, the problem becomes
\begin{equation}\label{eq:obj-multi-constraint}
    \begin{aligned}
         \max_{\theta} &~\mathbb{E}_{B \sim \mB} \left[  \mathbb{E}_{\alpha_{_B} \sim \pi(\cdot|B; \theta)} ~\left[R \left(\alpha_{\sss B} | B; w \right) \right] \right], \\
         &~\text{s.t. } ~c(\alpha_{\sss B}) \leq B, ~B \sim \mB,
     \end{aligned}
\end{equation}
where $\mathbb{E}_{B \sim \mB} \left[  \cdot \right]$ denotes the expectation over the distribution of budget. 
Unlike Eqn.~(\ref{eq:obj-single-constraint}), $\pi(\cdot|B;\theta)$ is the learned policy conditioned on the budget of $B$.
In practice, we use policy gradient to learn the architecture generator. 
To encourage exploration, we follow~\cite{pham2018efficient,guo2019nat,guo2021towards} to introduce an entropy regularization. Please refer to the supplementary materials for more details.

\textbf{Advantages over existing NAS methods.}
Our PNAG exhibits two advantages over existing NAS methods.
\emph{First}, our PNAG is able to share the learned knowledge across the search processes under different budgets, which greatly improves the search results (see Table~\ref{tab:pareto_learning}).
The main reason is that, once we find a good architecture for one budget, we may easily obtain a competitive architecture for a larger/smaller budget by slightly modifying some components (model width or kernel size).
\emph{Second}, given a well-trained PNAG, we can directly use it to generate feasible architectures for any required budget via inference, which is very efficient and practically useful (see Table~\ref{tab:generation_cost}).



\subsubsection{Vector Representation of Budget Bounds} \label{sec:rep_budget}

To learn the architecture generator, we still have to consider how to represent the budget bound $B$ as the inputs of \sexyname.
As mentioned before, our \sexyname considers $K$ discrete budgets during training. 
To represent different budgets, 
we use an embedding vector~\cite{pham2018efficient} to represent different budgets (See details in Section~\ref{sec:rep_budget}). 
Following~\cite{pham2018efficient}, we build a learnable embedding vector $\bb = g(B)$ for each sampled budget $B$. We incorporate these learnable embedding vectors into the parameters of the architecture generator and train them jointly. In this way, we are able to automatically learn the vectors of these budgets and encourage \sexyname to produce feasible architectures.  

As mentioned before, we only sample a set of discrete budgets to train \sexyname. To accommodate all the budgets belonging to a continuous space, we propose an embedding interpolation method to represent a budget with any possible value.
Specifically, we perform a linear interpolation between the embedding of two adjacent discrete budgets to represent the considered budgets.
For a target budget ${B}$ between two sampled budgets $B_1 {<} {B} {<} B_2$, the linear interpolation of the budget vector $\bb$ can be computed by
\begin{equation*}\label{eq:interpolation}
\begin{aligned}
  \bb = g({B}) = \xi  g(B_1) {+} (1 {-} \xi)  g(B_2),
  \text{~where~} \xi = \frac{{B_2} {-} B} {B_{2} {-} B_{1}},
\end{aligned}
\end{equation*}
Here, $\xi \in [0,1]$ denotes the weight of $B_1$ in interpolation.



\subsection{Learning the Architecture Evaluator $R(\cdot|B;w)$}\label{sec:reward}

Given diverse budgets, an architecture should have different rewards/scores regarding whether it satisfies the corresponding budget constraint.
However, it is non-trivial to manually design a reward function for each budget. Instead, we propose to learn an architecture evaluator to automatically predict the score.
To this end, we build an evaluator with three fully connected layers. Given any architecture $\beta$ and a budget $B$,  we seek to predict the performance $R(\beta|B;w)$ of $\beta$ under the budget $B$. 
Since we have no ground-truth labels for training, following~\cite{freund2003efficient,burges2005learning,chen2009ranking}, we learn the evaluator via pairwise architecture comparisons.

\begin{figure*}[t]
	\begin{minipage}[t]{0.47\linewidth}
	\centering
        \includegraphics[width=0.84\linewidth]{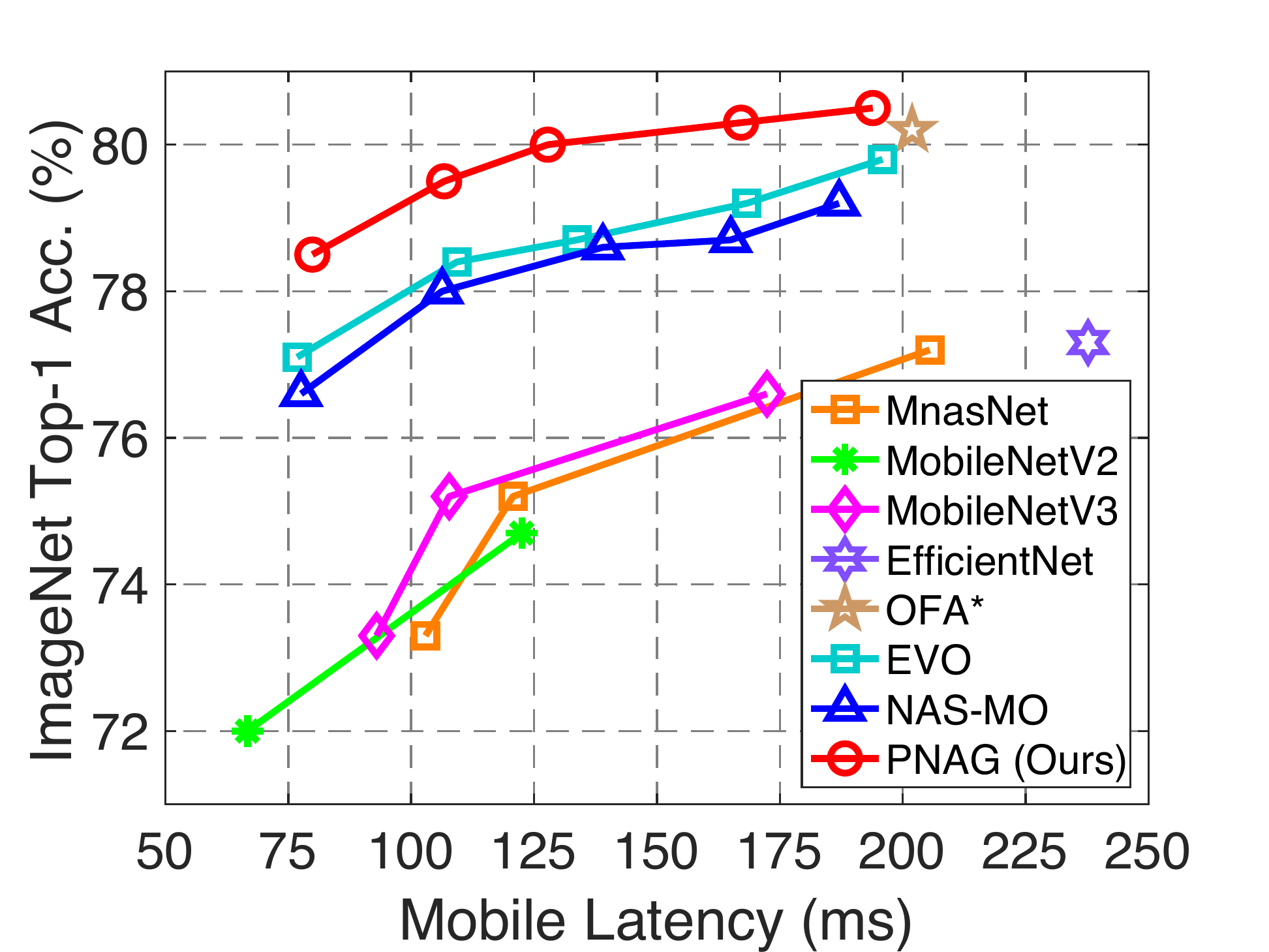}
        \caption{
        Comparisons of the architectures obtained by different methods on a mobile device (Qualcomm Snapdragon 821).
        }
        \label{fig:mobile_compare}
	\end{minipage}\hfill
	\begin{minipage}[t]{0.47\linewidth}
	\centering
        \includegraphics[width=0.84\linewidth]{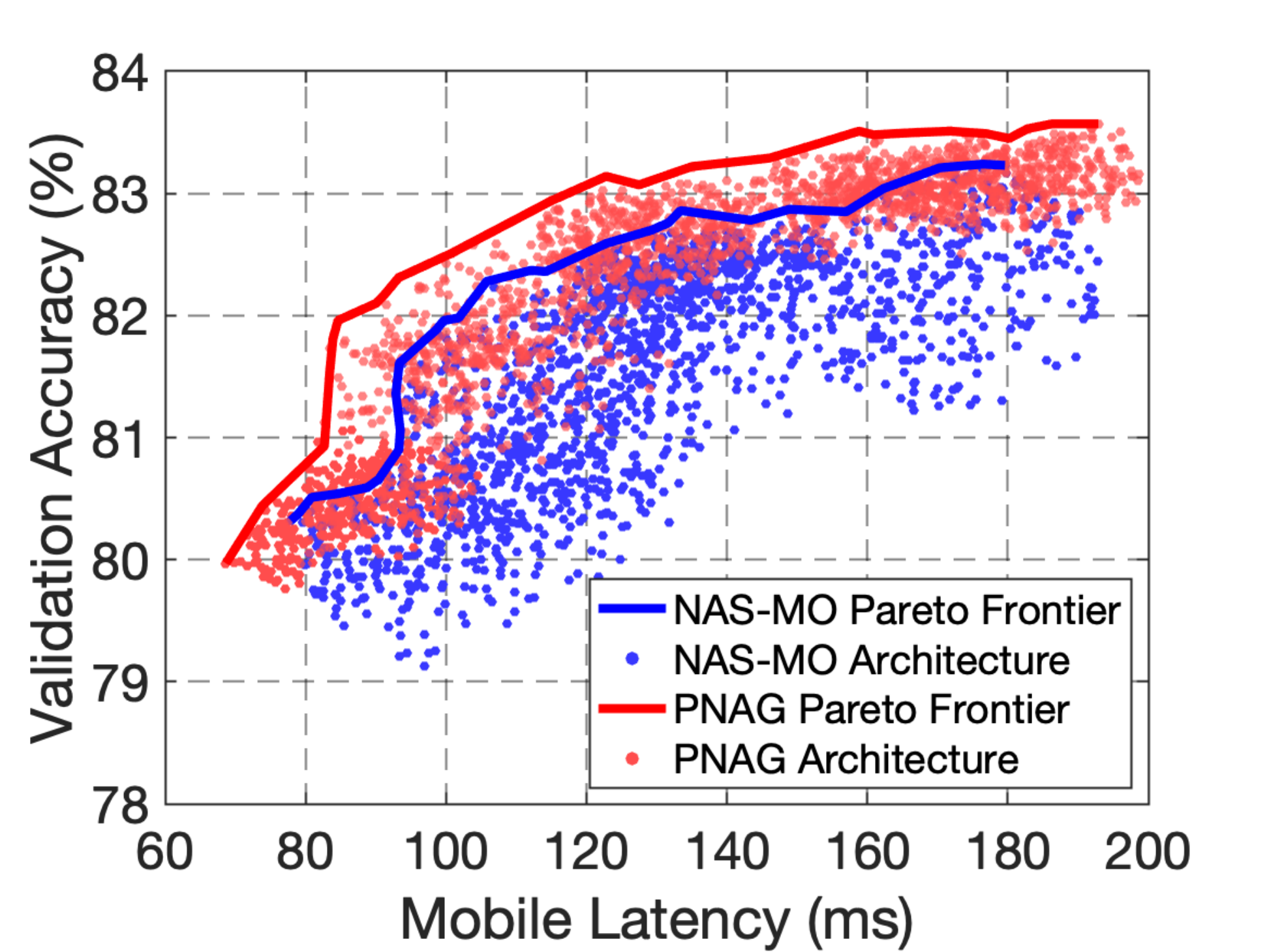}
        \caption{
        Comparisons of the Pareto frontiers of the {generated} architectures between NAS-MO and PNAG. Here, we report the accuracy evaluated on the constructed validation set.
        }
        \label{fig:pareto_curve}
	\end{minipage}
\end{figure*}

\begin{figure*}[t]
	\centering
	\subfigure[Ground-truth latency histogram.]{
		\includegraphics[width = 0.67\columnwidth]{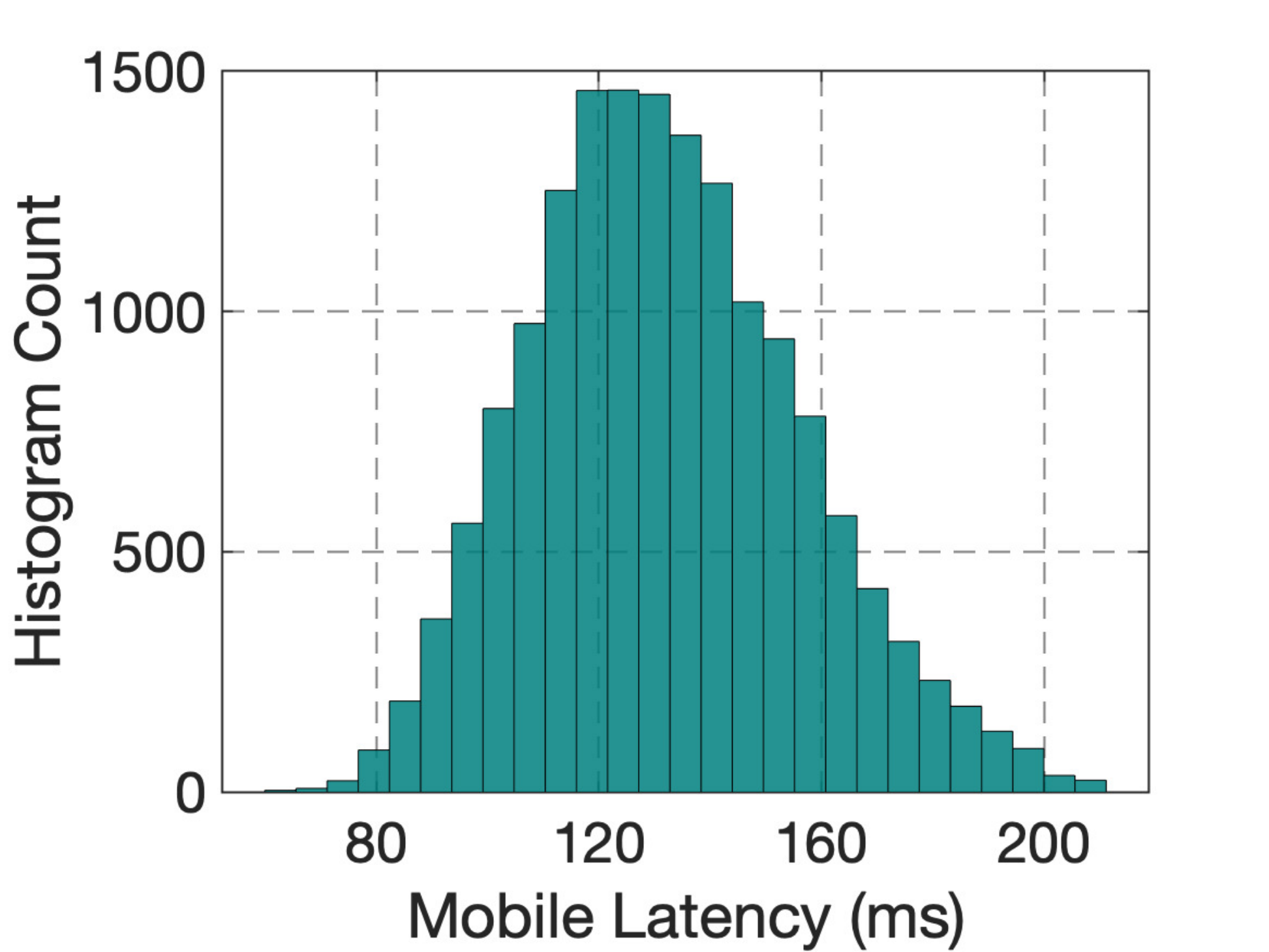}\label{fig:mobile_dist}
	}~
	\subfigure[{Generation results with $B{=}110$ms.}]{
		\includegraphics[width = 0.67\columnwidth]{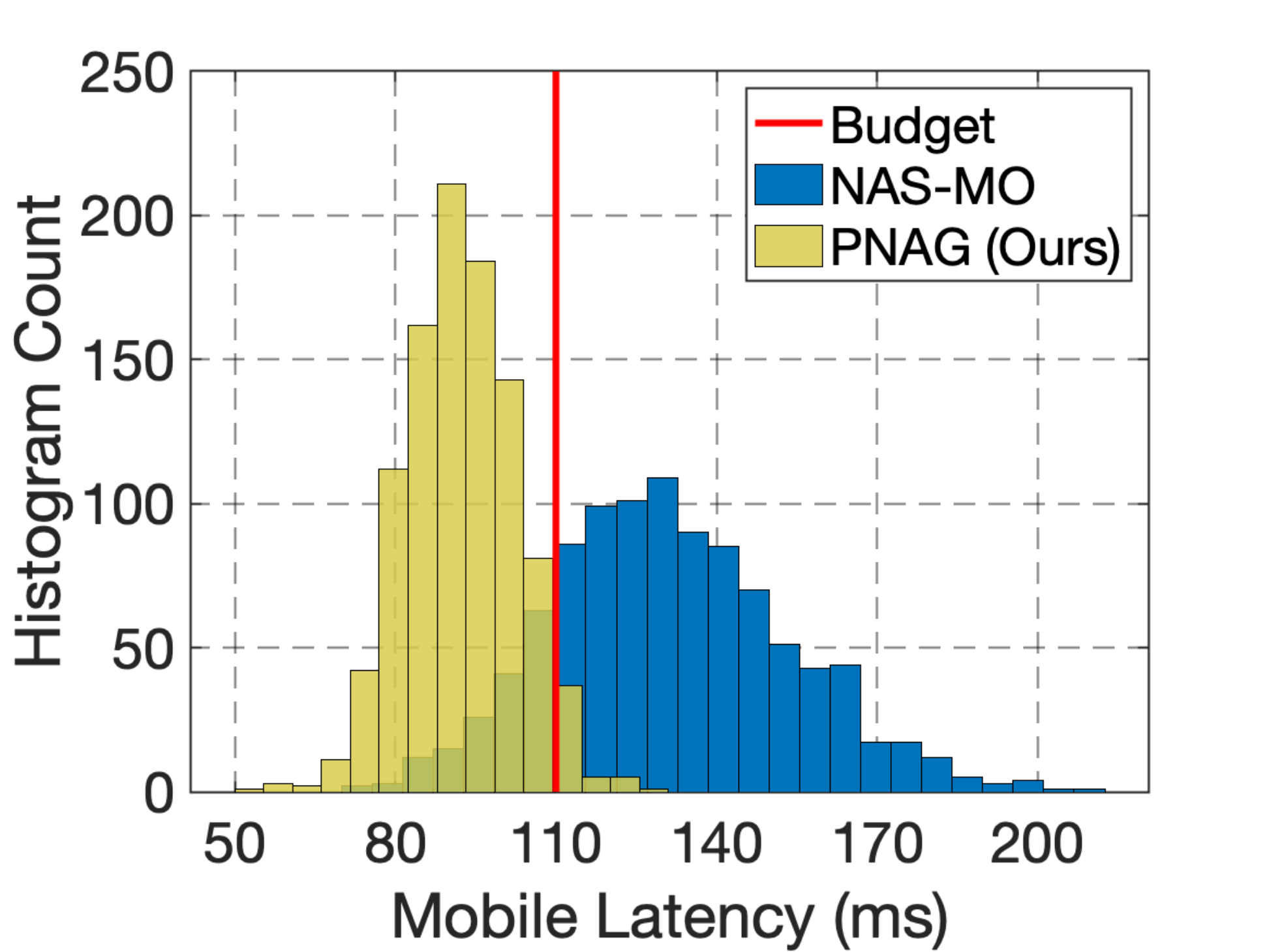}\label{fig:histogram_110}
	}~
	\subfigure[Generation results with $B{=}140$ms.]{
		\includegraphics[width = 0.67\columnwidth]{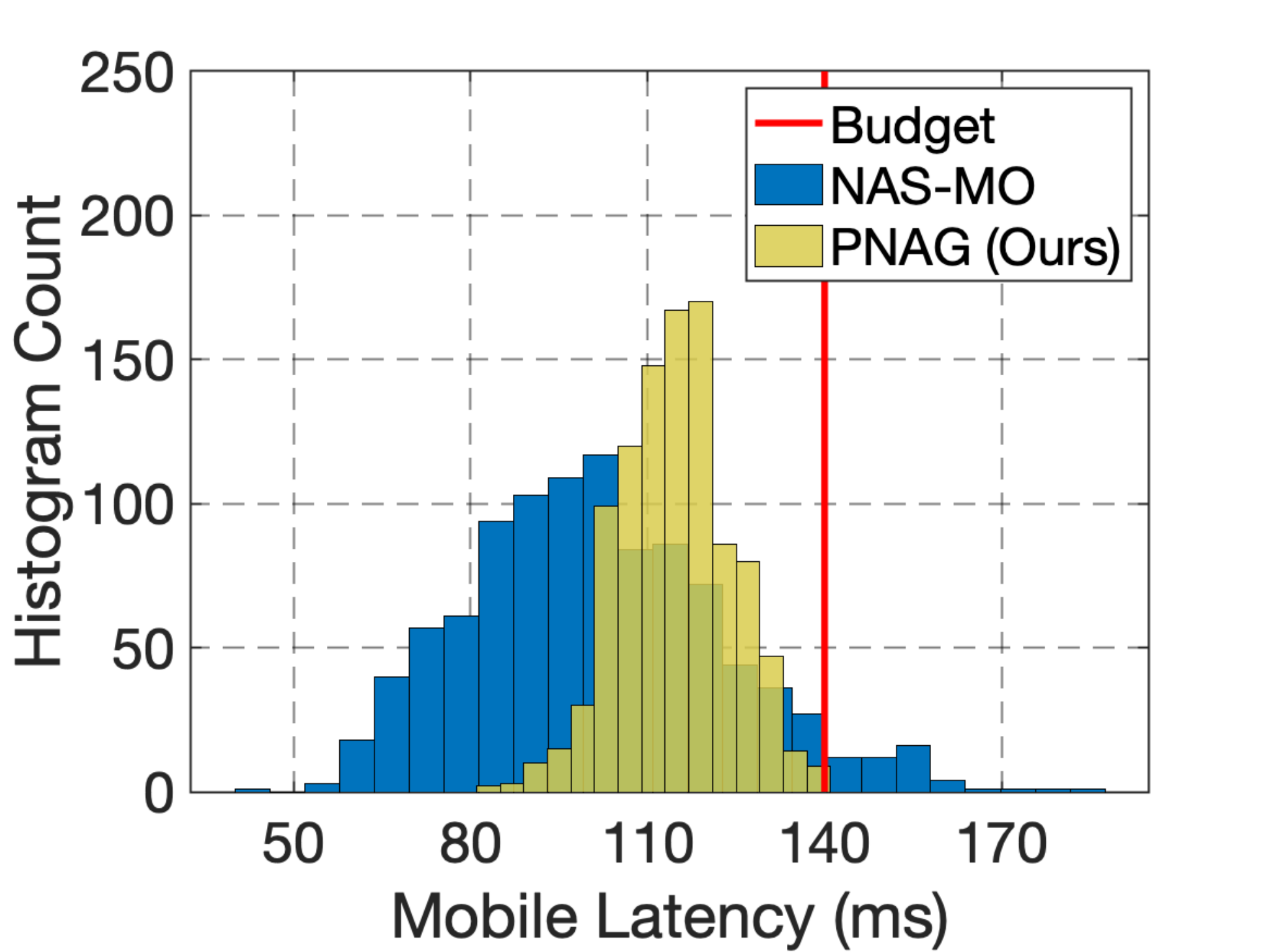}\label{fig:histogram_140}
	}
	\caption{
    Latency histograms {of sampled architectures} on mobile devices. (a) Ground-truth latency histogram of $16,000$ architectures that are uniformly sampled from the search space. (b) The latency histogram of $1,000$ architectures sampled by different methods given $B{=}110$ms. {(c) The latency histogram of $1,000$ architectures sampled by different methods given $B{=}140$ms.}}
	\label{fig:distribution}
\end{figure*}

\subsubsection{Training Method of $R(\cdot|B;w)$}
To obtain a promising evaluator, we train the architecture evaluator using a pairwise ranking loss, which has been widely used in ranking problems~\cite{freund2003efficient,burges2005learning,chen2009ranking}.
Specifically, we collect $M$ architectures with accuracy and latency, and record them as a set of triplets $\{(\beta_i, c(\beta_i), {\rm Acc}(\beta_i))\}_{i=1}^{M}$.
Thus, given $M$ architectures, we have $M(M{-}1)$ architecture pairs $\{(\beta_{i}, \beta_{j})\}$ in total after omitting the pairs with themselves.
Assuming that we have $K$ budgets, the pairwise ranking loss becomes
\begin{equation}\label{eq:ranking_loss}
\begin{aligned}
    L(w) = &\frac{1}{KM(M{-}1)} \sum_{k=1}^K \sum_{i=1}^{M} \sum_{j=1, j \neq i}^{M} \phi \Big( d( \beta_{i}, \beta_{j}, B_k)    \\
     &\cdot \big[ R(\beta_{i}|B_k;w) - R(\beta_{j}|B_k;w) \big]  \Big),
\end{aligned}
\end{equation}
where $d\big(\beta_1, \beta_2, B_k \big)$ denotes a function to indicate whether $\beta_{i}$ is better than $\beta_{j}$ under the budget $B_k$, as will be discussed in Section~\ref{sec:pareto_dominance}. $\phi(z) = \max (0, 1-z)$ is a hinge loss function and we use it to enforce the predicted ranking results $R(\beta_{i}|B_k;w) - R(\beta_{j}|B_k;w)$ to be consistent with the results of $d( \beta_{i}, \beta_{j}, B_k)$ obtained by a comparison rule based on Pareto dominance.

\subsubsection{Pareto Dominance Rule}\label{sec:pareto_dominance}
To compare the performance between two architectures, we need to define a reasonable function $d\big(\beta_1, \beta_2, B \big)$ in Eqn.~(\ref{eq:ranking_loss}). To this end, we define a Pareto dominance to guide the design of this function.
Specifically, Pareto dominance requires that the quality of an architecture should depend on both the satisfaction of budget and accuracy.
That means, given a specific budget $B$, a good architecture should be the one with the cost lower than or equal to $B$ and with high accuracy.
In this sense, we use Pareto dominance to compare two architectures and judge which one is dominative.

Given any two architectures $\beta_1, \beta_2$, if both of them satisfy the budget constraints (namely $c(\beta_1) \leq B$ and $c(\beta_2) \leq B$),  then $\beta_1$ dominates  $\beta_2$ if ${\rm Acc}(\beta_1) \geq {\rm Acc}(\beta_2)$. 
Moreover, when at least one of $\beta_1, \beta_2$ violates the budget constraint, clearly we have that $\beta_1$ dominates $\beta_2$ if $c(\beta_1) \leq c(\beta_2)$.
Formally, we define the Pareto dominance function $d\big(\beta_1, \beta_2, B \big)$ to reflect the above rules:

\begin{equation}\label{eq:compare_rule}
    d\big(\beta_1, \beta_2, B \big) = 
    \begin{cases} 
        ~1, ~~~~~{\rm if} ~~\left(c(\beta_1) \leq B ~{\land}~ c(\beta_2) \leq B \right) \\
        ~~~~~~~~~~~~~{\land}~ (\textcolor{black}{{\rm Acc}(\beta_1) \geq {\rm Acc}(\beta_2)}); \vspace{3 pt} \\ 
               -1, ~~~{\rm else~if} ~~\left(c(\beta_1) \leq B ~{\land}~ c(\beta_2) \leq B \right) \\
        ~~~~~~~~~~~~~{\land}~ (\textcolor{black}{{\rm Acc}(\beta_1) < {\rm Acc}(\beta_2)}); \vspace{3 pt} \\ 
        ~1, ~~~~~{\rm else~if} ~~c(\beta_1) \leq ~ c(\beta_2); \vspace{3 pt} \\ 
        -1, ~~~{\rm otherwise}.
    \end{cases}
\end{equation}
Based on Eqn.~(\ref{eq:compare_rule}), we have $d(\beta_1, \beta_2, B) = - d(\beta_2, \beta_1, B)$ if $\beta_1 \neq \beta_2$, making it a symmetric metric \wrt $\beta_1$ and $\beta_2$. 

\begin{remark}
The accuracy constraint ${\rm Acc}(\beta_1) \geq {\rm Acc}(\beta_2)$ plays an important role in the proposed Pareto dominance function $d\big(\beta_1, \beta_2, B \big)$. Without the accuracy constraint, we may easily find the architectures with very low computation cost and poor performance (See results in Table~\ref{tab:diff_reward}).
\end{remark}

\begin{table*}[t!] 
    \centering
    \caption{
    Comparisons with state-of-the-art architectures on mobile devices. $^*$ denotes the best architecture reported in the original paper. 
    ``-'' denotes the results that are not reported. All the models are evaluated on $224 \times 224$ images of ImageNet.
    }
    \resizebox{0.85\textwidth}{!}
    {               
        \begin{tabular}{ccccccc}
        \toprule [0.15em]
        \multirow{2}[0]{*}{Architecture}  & \multirow{2}[0]{*}{Latency (ms)}  & \multicolumn{2}{c}{Test Accuracy (\%)} & \multirow{2}[0]{*}{\#Params (M)}  & \multirow{2}[0]{*}{\#MAdds (M)}   & Search Cost  \\
        \cline{3-4}
        & & Top-1 & Top-5 & & & (GPU Days) \\
        \midrule [0.1em]
        MobileNetV3-Small (1.0$\times$)~\cite{howard2019searching} & 39.8 & 67.4 & - & 2.4 & 56  \\
        MobileNetV3-Large (0.75$\times$)~\cite{howard2019searching} & 93.0 & 73.3 & - & 4.0 & 155 & -  \\
        MobileNetV2 (1.0$\times$)~\cite{sandler2018mobilenetv2} & 90.3 & 72.0 & - & 3.4 & 300 & - \\
        FBNetV2~\cite{wan2020fbnetv2} & - & 76.3 & 92.9 & - & 321 & 30.0 \\
        MnasNet-A1 (0.5$\times$)~\cite{tan2019mnasnet} & 37.5 & 68.9 & 88.4 & 2.1 & 105 \\
        SGNAS-B~\cite{huang2021searching} & - & 76.8 & - & - & 326 & 0.3 \\
        {EVO}-80 & 76.8 & 77.1 & 93.3 & 6.1	& 350 & 0.7 \\
        NAS-MO-80 & 77.6 & 76.6 & 93.2 & 7.9 & 340 & 0.7 \\
        \sexyname-80 & 79.9 & \textbf{78.3} & \textbf{94.0}  & 7.3 & 349 & 0.7 \\
        \midrule
        FBNet-A~\cite{wu2019fbnet} & 91.7 & 73.0 & - & 4.3 & 249 & 9.0 \\
        SGNAS-A~\cite{huang2021searching} & - & 77.1 & - & - & 373 & 0.3 \\
        ProxylessNAS-Mobile~\cite{cai2018proxylessnas} & 97.3 & 74.6 & - & 4.1 & 319 & 8.3 \\
        ProxylessNAS-CPU~\cite{cai2018proxylessnas} & 98.5 & 75.3 & - & 4.4 & 438 & 8.3 \\
        MobileNetV3-Large (1.0$\times$)~\cite{howard2019searching} & 107.7 & 75.2 & - & 5.4 & 219 & - \\
        {EVO}-110 & 109.3 & 78.4 & 94.0 & 10.2 & 482 & 0.7 \\
        NAS-MO-110 & 106.3 & 78.0 & 93.8 & 8.4 & 478 & 0.7 \\
        \sexyname-110 & 106.8 & \textbf{79.4} & \textbf{94.5} & 9.9 & 451 & 0.7 \\
        \midrule
        RandWire~\cite{xie2019exploring} & - & 74.7 & 92.2 & 5.6 & 583 & - \\
        ProxylessNAS-GPU~\cite{cai2018proxylessnas} & 123.3 & 75.1 & - & 7.1 & 463 & 8.3  \\
        MnasNet-A1 (1.0$\times$)~\cite{tan2019mnasnet} & 120.7  & 75.2 & 92.5 & 3.4 & 300 & $\sim$3792  \\
        FBNet-C~\cite{wu2019fbnet} & 135.2 & 74.9 & - & 5.5 & 375 & 9.0 \\
        {EVO}-140 & 133.7 & 78.7 & 94.1 & 9.1 & 488 & 0.7 \\
        NAS-MO-140 & 139.0 & 78.6 & 94.0 & 9.5 & 486 & 0.7 \\
        \sexyname-140 & 127.8 & \textbf{79.8} & \textbf{94.7}  & 9.2 & 492 & 0.7 \\
        \midrule
        NSGANetV1~\cite{lu2020multi} & - & 76.2 & 93.0 & 5.0 & 585 & 27 \\
        PONAS-C~\cite{huang2020ponas} & 145.1 & 75.2 & - & 5.6 & 376 & 8.8 \\
        P-DARTS~\cite{chen2019progressive} & 168.7 & 75.6 & 92.6 & 4.9 & 577 & 3.8 \\ 
        BigNAS-L~\cite{yu2020bignas} & - & 79.5 & - & 6.4 & 586 & 1.5 \\ 
        {EVO}-170 & 168.3 & 79.2 & 94.4 & 10.7 & 661 & 0.7 \\
        NAS-MO-170 & 165.0 & 78.7 & 94.4 & 8.5 & 584 & 0.7 \\
        \sexyname-170 & 167.1 & \textbf{80.3} & \textbf{95.0} & 10.0 & 606 & 0.7 \\
        \midrule
        NSGANetV2~\cite{lu2020nsganetv2} & - & 79.1 & 94.5 & 8.0 & 400 & 1 \\
        NAGO~\cite{ru2020neural} & - & 76.8 & 93.4 & 5.7 & - & 20.0 \\
        PC-DARTS~\cite{xu2020pcdarts} & 194.1 & 75.8 & 92.7 & 5.3 & 597 & 0.1 \\
        EfficientNet B0~\cite{EfficientNet} & 237.7 & 77.3 & 93.5 & 5.3 & 390 & - \\
        Cream-L~\cite{peng2020cream} & - & 80.0 & 94.7 & 9.7 & 604 & 12 \\
        OFA$^*$~\cite{cai2019once} & 201.9 & 80.2 & 95.1 & 9.1 & 743 & 51.7 \\
        {EVO}-200 & 195.9 & 79.8  & 94.5 & 11.0 & 783 & 0.7 \\
        NAS-MO-200 & 187.4 & 79.2  & 94.4 & 9.1 & 630 & 0.7 \\
        \sexyname-200 & 193.9 & \textbf{80.5} & \textbf{95.2}  & 10.4 & 724 & 0.7 \\
        \bottomrule[0.15em]
        \end{tabular}  
    }

    \label{tab:mobile_comp}     
\end{table*}

\section{Experiments}\label{sec:exp}

We apply the proposed \sexyname to produce architectures under diverse latency budgets evaluated on three hardware platforms, including a mobile device (equipped with a Qualcomm Snapdragon 821 processor), a CPU processor (Intel Core i5-7400), and a GPU card (NVIDIA TITAN X).
For convenience, we use ``Architecture-$B$'' to represent the {generated} architecture that satisfies the latency budget w.r.t. $B$, \eg, \sexyname-80. Our code and all the pretrained models are available at \href{https://github.com/guoyongcs/PNAG}{https://github.com/guoyongcs/PNAG}.

\subsection{Implementation Details}\label{sec:implementation}

~\indent \textbf{Search space}.
Following~\cite{cai2019once}, we use MobileNetV3~\cite{howard2019searching} as the backbone to build the search space~\cite{cai2019once,huang2020ponas}. 
We divide a network into several units.
To find promising architectures, we allow each unit to have {1)} any {numbers of layers} (\ie, depth) chosen from $\{2,3,4\}$, {2)} any {width expansion ratios in each layer} (\ie, width) chosen from $\{3,4,6\}$, and {3)} any {kernel sizes} chosen from $\{3,5,7\}$. 
We build the model with 5 units{.} Thus, there are $3 {\times} 3$ combinations of widths and kernel sizes for each layer. 

\textbf{Training the supernet}.
To accelerate the training of supernet, we follow~\cite{wu2019fbnet} to randomly choose 100 classes from original 1000 classes in ImageNet for training and train the supernet with progressive shrinking strategy~\cite{cai2019once} for 90 epochs.
We treat 80\% of these data as the training set to train the supernet and the rest 20\% as the validation set to measure the validation accuracy of candidate architectures (we report such validation accuracy in Figs.~\ref{fig:search_direction} and~\ref{fig:pareto_curve}).
We consider the original ImageNet validation set as the test data and report the test accuracy of candidate architectures on them in all the other tables and figures.
Based on a NVIDIA V100 GPU, the training process of the supernet takes around \emph{15 GPU hours} (\ie, 0.6 GPU days).



\textbf{Training the architecture evaluator.} 
We collect {$M{=}16,000$} architectures by uniformly sampling architectures from the search space $\Omega$ (See Fig.~\ref{fig:mobile_dist}) following~\cite{cai2019once} and obtain the latency ranges on three hardware devices.
We deploy these architectures to different devices and measure the latency over a batch of images.
Specifically, we measure the latency on mobile and CPU devices with a batch size of 1. Since the inference on GPU is too fast to obtain the accurate latency, we measure the latency with a batch size of 64 on NVIDIA TITAN X.
We compute the accuracy ${\rm Acc}(\cdot)$ on our validation set (\ie, 20\% samples of 100 selected classes in ImageNet).
We train the architecture evaluator for $250$ epochs. 
The learning rate is initialized to $0.1$ and decreased to $1 {\times} 10^{-3}$ with a cosine annealing. 
Following~\cite{cai2019once}, we train two predictors to predict the latency and validation accuracy, respectively.
We set the dimension of the embedding vector of budgets to 64.
We emphasize that training the architecture evaluator is very efficient and only takes less than \emph{0.2 GPU hours}.

\textbf{Training the architecture generator.} 
We train the model for $120$k iterations using an Adam optimizer with a learning rate of $3\times10^{-4}$.
Following ENAS~\cite{pham2018efficient}, we sample $N{=}1$ architecture at each iteration and find it works well in practice.
We select $K{=}10$ latency budgets by evenly dividing the range.
We add an entropy regularization term to the reward weighted by $1 {\times} 10^{-3}$. 
Note that training the architecture generator approximately takes \emph{2 GPU hours}.
When evaluating the searched architectures, following~\cite{cai2019once,lu2021neural}, we first obtain the parameters from the OFA full network and then finetune them for 75 epochs to obtain the final performance. 

\begin{figure*}[t]
	\begin{minipage}[t]{0.47\linewidth}
	\centering
        \includegraphics[width = 0.84\columnwidth]{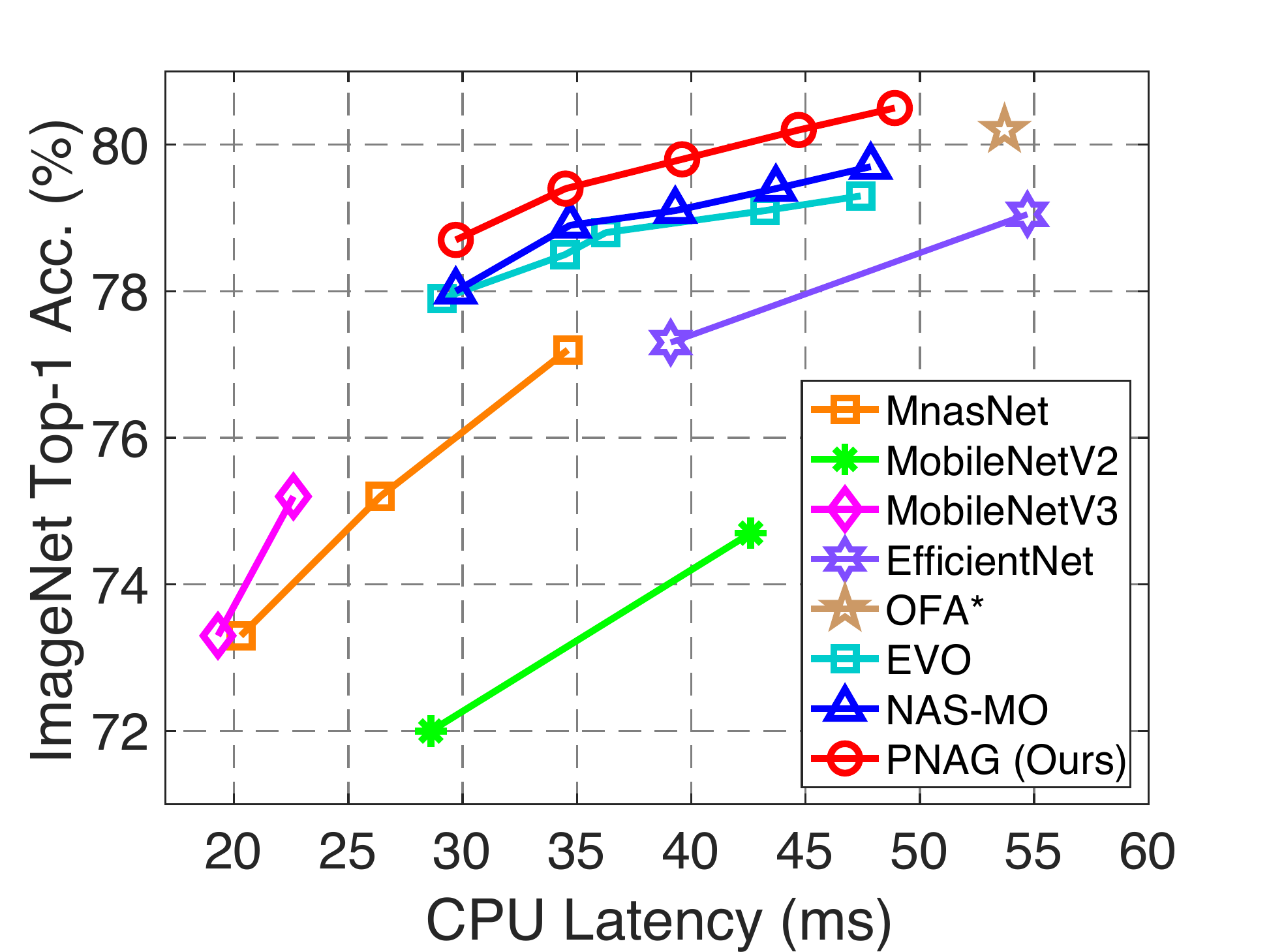}\label{fig:cpu_compare}
        \caption{
        Comparisons of the architectures obtained by different methods on a Core i5-7400 CPU.
        }
        \label{fig:cpu_compare}
	\end{minipage}\hfill
	\begin{minipage}[t]{0.47\linewidth}
	\centering
        \includegraphics[width = 0.84\columnwidth]{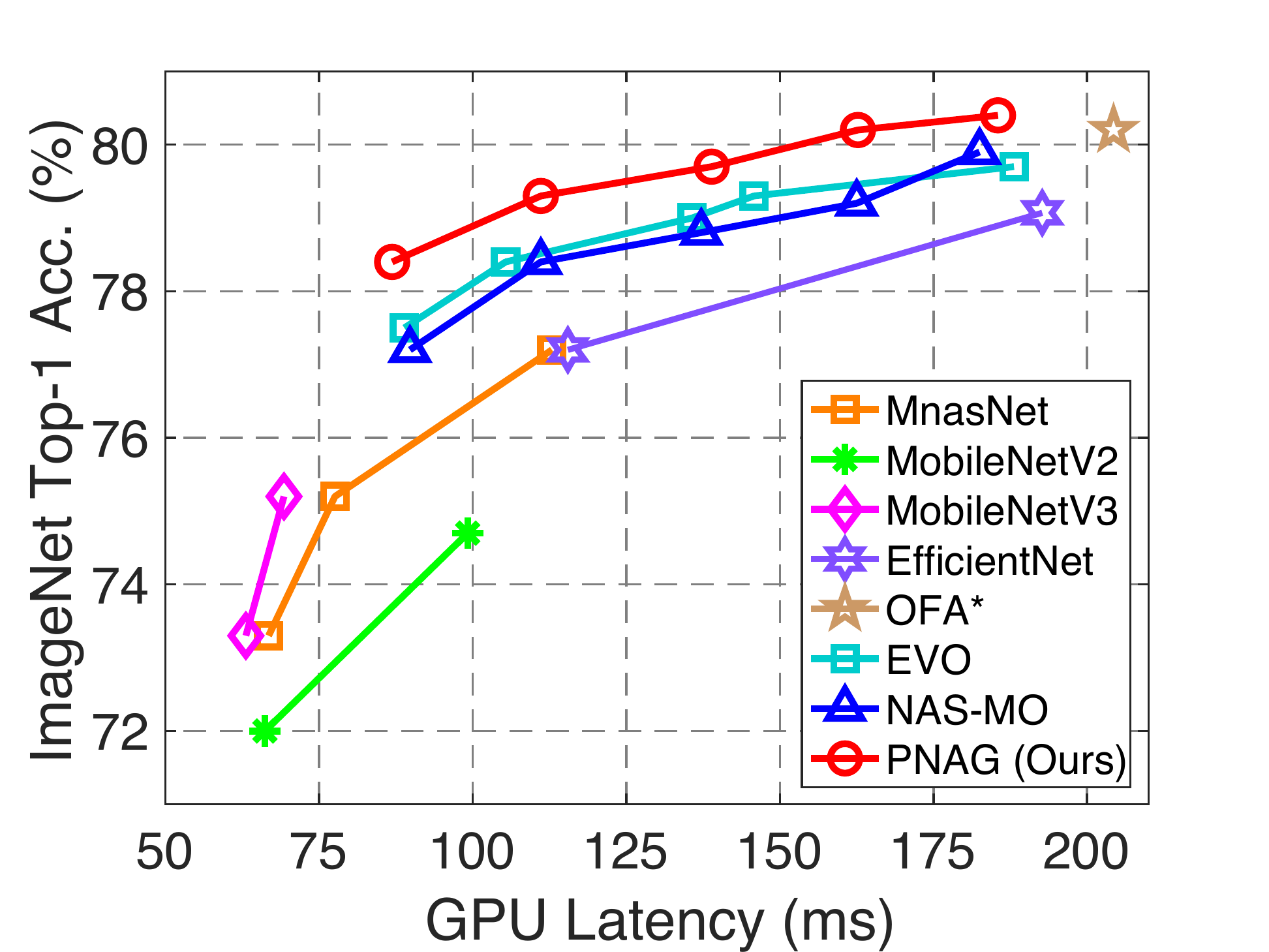}\label{fig:gpu_compare}
        \caption{
        Comparisons of the architectures obtained by different methods on a  NVIDIA TITAN X GPU.
        }
        \label{fig:gpu_compare}
	\end{minipage}
\end{figure*}

\subsection{Compared Methods}

{To investigate the effectiveness of the proposed method, we compare our PNAG with two variants:
}
1) \textbf{EVO} uses the evolutionary search method~\cite{real2019regularized} to perform architecture search.
2) \textbf{NAS-MO} conducts architecture search based by exploiting the multi-objective reward~\cite{tan2019mnasnet}.
{We also compare our method with several state-of-the-art methods, including ENAS~\cite{pham2018efficient}, DARTS~\cite{liu2018darts}, P-DARTS~\cite{chen2019progressive}, PC-DARTS~\cite{xu2020pcdarts}, MNasNet~\cite{tan2019mnasnet}, MobileNetV2~\cite{sandler2018mobilenetv2}, MobileNetV3~\cite{howard2019searching}, FBNet~\cite{wu2019fbnet}, FBNetV2~\cite{wan2020fbnetv2}, ProxylessNAS~\cite{cai2018proxylessnas}, EfficientNet~\cite{EfficientNet}, OFA~\cite{real2019regularized}, Cream~\cite{peng2020cream}, PONAS~\cite{huang2020ponas}, BigNAS~\cite{yu2020bignas}, and NSGANetV2~\cite{lu2020nsganetv2}.}

\subsection{Architecture Search for Mobile Devices}

In this experiment, we train our PNAG to produce feasible architectures for the latency budgets based on a mobile device (Qualcomm Snapdragon
821 processor). 
Based on the proposed budget interpolation method in Section~\ref{sec:rep_budget}, our PNAG is flexible to generate feasible architectures for any arbitrary budget. 
To evaluate our method, for simplicity, we manually choose 5 latency budgets $\{$80ms, 110ms, 140ms, 170ms, 200ms$\}$ and reports the results under each of them. The other budgets are also possible.

We compare our \sexyname with state-of-the-art methods given different latency budgets evaluated on he considered mobile device. 
In Fig.~\ref{fig:mobile_compare}, we compare the architectures searched by different methods in terms of both accuracy and latency. We draw the following conclusions. \emph{First}, our PNAG (red line) consistently generates better architectures than the considered variants EVO and NAS-MO under diverse budgets. \emph{Second}, our best architecture (the rightmost point of the red line) yields a better trade-off between accuracy and latency than a strong baseline OFA$^*$, \ie, the best architecture reported in~\cite{cai2019once}.
For convenience, we put more detailed comparison results in Table~\ref{tab:mobile_comp}. 
Given diverse latency budgets, our PNAG greatly outperforms the compared NAS methods in terms of the accuracy of the generated/searched architectures. Specifically, our PNAG-200 yields the best accuracy of 80.5, which is better than the best reported results in OFA~\cite{cai2019once}, namely OFA$^*$.
We also highlight that, besides the superior performance, the total training cost of our PNAG is about 0.7 GPU days\footnote{We report the training cost of each component of PNAG in Section~\ref{sec:implementation}}, which is much more efficient than most state-of-the-art NAS methods, such as~\cite{cai2019once,peng2020cream,yu2020bignas}.

Moreover, we compare the learned/searched frontiers of different methods
{and show the comparisons of Pareto frontiers in Fig.~\ref{fig:pareto_curve}.
We plot all the architectures produced by different methods to form the Pareto frontier. 
Specifically, we use the architectures searched by multiple independent runs under different budgets for NAS-MO.
For \sexyname, we use linear interpolation to generate architectures that satisfy different budgets.}
From Fig.~\ref{fig:pareto_curve}, our \sexyname finds a better frontier than NAS-MO due to the shared knowledge across the search process under different budgets.
We also visualize the latency histograms of the architectures evaluated on mobile devices in Fig.~\ref{fig:histogram_110} and {Fig.~\ref{fig:histogram_140}}.
{Given latency budgets of $110$ms and $140$ms, NAS-MO is prone to produce a large number of architectures that cannot satisfy the target budgets. 
These results show that it is hard to design the multi-objective reward to obtain the preferred architectures.}
Instead, \sexyname uses the Pareto dominance reward to encourage the architectures to satisfy the desired budget constraints. 
In this sense, most architectures generated by our \sexyname are able to fulfill the target budgets.
We put more visual results of latency histograms \wrt other latency budgets in the supplementary.

\begin{table*}[t!]   
    \caption{
    Comparisons with state-of-the-art architectures on Intel Core i5-7400 CPU. $^*$ denotes the best architecture reported in the original paper. 
    ``-'' denotes the results that are not reported. All the models are evaluated on $224 \times 224$ images of ImageNet.
    }
    \centering
    \resizebox{0.80\textwidth}{!}
    {               
        \begin{tabular}{ccccccc}
        \toprule [0.15em]

        \multirow{2}[0]{*}{Architecture}  & \multirow{2}[0]{*}{Latency (ms)}  & \multicolumn{2}{c}{Test Accuracy (\%)} & \multirow{2}[0]{*}{\#Params (M)}  & \multirow{2}[0]{*}{\#MAdds (M)}   & Search Cost  \\
        \cline{3-4}
        & & Top-1 & Top-5 & & & (GPU Days) \\
        \midrule [0.1em]
        MobileNetV2 (1.0$\times$)~\cite{sandler2018mobilenetv2} & 28.6 & 72.0 & - & 3.4 & 300 & - \\
        MobileNetV3-Large (1.0$\times$)~\cite{howard2019searching} &  22.6 & 75.2 & - & 5.4 & 219 & - \\
        FBNet-C~\cite{wu2019fbnet} & 25.7 & 74.9 & - & 5.5 & 375 & 9.0 \\
        SGNAS-B~\cite{huang2021searching} & - & 76.8 & - & - & 326 & 0.3 \\
        EVO-30 & 29.1 & 77.9 & 93.8 & 7.9 & 385 & 0.7 \\
        NAS-MO-30 & 29.7 & 77.5 & 93.7 & 6.6 & 353 & 0.7 \\
        \sexyname-30 (Ours) & 29.7 & \textbf{78.3} & \textbf{94.1} & 7.6 & 335 & 0.7 \\
        \midrule
        ProxylessNAS-CPU~\cite{cai2018proxylessnas}  & 34.6 & 75.3 & - & 4.4 & 438 & 8.3 \\
        MnasNet-A1 (1.4$\times$)~\cite{tan2019mnasnet}  & 34.6 & 77.2 & 93.5 & 6.1 & 592 & $\sim$3792 \\
        EVO-35 & 34.5 & 78.5 & 94.3 & 8.2 & 354 & 0.7 \\
        NAS-MO-35 & 34.7 & 78.3 & 94.0 & 7.9 & 478 & 0.7 \\
        \sexyname-35 (Ours) & 34.5 & \textbf{79.4} &  \textbf{94.5} & 8.4 & 431 & 0.7 \\
        \midrule
        ResNet-18~\cite{he2016deep} & 38.6 & 69.8 & 90.1 & 11.7 & 1814 & - \\
        EfficientNet B0~\cite{EfficientNet}  & 39.1 & 77.3 & 93.5 & 5.3 & 390 & - \\
        EVO-40 & 36.3 & 78.8 & 94.6 & 8.4 & 388 & 0.7 \\
        NAS-MO-40 & 39.3 & 78.6 & 94.3 & 8.3 & 491 & 0.7 \\
        \sexyname-40 (Ours) & 39.6 & \textbf{79.8} & \textbf{94.9} & 9.4 & 502 & 0.7 \\       
        \midrule
        MobileNetV2 (1.4$\times$)~\cite{sandler2018mobilenetv2}  & 42.6 & 74.7 & - & 6.9 & 585 & - \\
        EVO-45 & 43.2 & 79.1 & 94.6 & 9.1 & 481 & 51.7 \\
        NAS-MO-45 & 43.7 & 78.8 & 94.4 & 9.3 & 626 & 0.7 \\
        \sexyname-45 (Ours) & 44.7 & \textbf{80.2} & \textbf{95.0} & 10.4 & 620 & 0.7 \\
        \midrule
        PONAS-C~\cite{huang2020ponas} & 52.2 & 75.2 & - & 5.6 & 376 & 8.8 \\
        OFA$^*$~\cite{cai2019once} & 53.7 & 80.2 & 95.1 & 9.1 & 743 & 51.7 \\
        EVO-50 & 47.4 & 79.3 & 94.7 & 9.1 & 511 & 0.7 \\
        NAS-MO-50 & 46.7 & 78.9 & 94.4 & 9.1 & 632 & 0.7 \\
        \sexyname-50 (Ours) & 48.9 & \textbf{80.5} & \textbf{95.1} & 10.5 & 682 & 0.7 \\
        \bottomrule[0.15em]
        \end{tabular}  
    }
    \label{tab:cpu_comp}     
\end{table*}

\begin{table*}[t!]
    \caption{
    Comparisons with state-of-the-art architectures on NVIDIA TITAN X GPU. $^*$ denotes the best architecture reported in the original paper. 
    ``-'' denotes the results that are not reported. All the models are evaluated on $224 \times 224$ images of ImageNet. 
    }
    \centering
    \resizebox{0.8\textwidth}{!}
    {               
        \begin{tabular}{ccccccc}
        \toprule [0.15em]

        \multirow{2}[0]{*}{Architecture}  & \multirow{2}[0]{*}{Latency (ms)}  & \multicolumn{2}{c}{Test Accuracy (\%)} & \multirow{2}[0]{*}{\#Params (M)}  & \multirow{2}[0]{*}{\#MAdds (M)}   & Search Cost  \\
        \cline{3-4}
        & & Top-1 & Top-5 & & & (GPU Days) \\
        \midrule [0.1em]
        
        ProxylessNAS-GPU~\cite{cai2018proxylessnas} & 84.7 & 75.1 & - & 7.1 & 463 & 8.3 \\
        MobileNetV2 (1.0$\times$)~\cite{sandler2018mobilenetv2} & 71.6 & 72.0 & - & 3.4 & 300 & - \\
        NAGO~\cite{ru2020neural} & - & 76.8 & 93.4 & 5.7 & - & 20.0 \\
        {EVO}-90 & 88.9 & 77.3 & 93.1 & 5.9 & 332 & 0.7 \\
        NAS-MO-90 & 89.8 & 75.4 & 92.4 & 4.9 & 266 & 0.7 \\
        \sexyname-90 (Ours) & 86.9 & \textbf{78.3} & \textbf{94.0} & 5.7 & 310 & 0.7 \\
        \midrule
        MnasNet-A1 (1.4$\times$)~\cite{tan2019mnasnet} & 112.9 & 77.2 & 93.5 & 6.1 & 592 & $\sim$3792 \\
        EfficientNet B0~\cite{EfficientNet} & 115.5 & 77.3 & 93.5 & 5.3 & 390 & - \\
        ENAS~\cite{pham2018efficient} & 110.8 & 73.8 & 91.7 & 5.6 & 607 & 0.5 \\
        {EVO}-115 & 105.4 & 78.4 & 94.1 & 8.4 & 388 & 51.7 \\
        NAS-MO-115 & 111.2 & 78.1 & 94.0 & 8.8 & 431 & 0.7 \\
        \sexyname-115 (Ours) & 111.2 & \textbf{79.3} & \textbf{94.6} & 8.9 & 411 & 0.7 \\
        \midrule
        {EVO}-140 & 135.7 & 78.9 & 94.4 & 9.1 & 481 & 0.7 \\
        NAS-MO-140 & 137.2 & 78.4 & 94.1 & 8.8 & 470 & 0.7 \\
        \sexyname-140 (Ours) & 138.9 & \textbf{79.7} & \textbf{94.9} & 9.7 & 510 & 0.7 \\
        \midrule
        ResNet-50~\cite{he2016deep} & 159.8 & 76.2 & 92.9 & 25.6 & 4087 & - \\
        {EVO}-165 & 164.1 & 79.1 & 94.5 & 10.7 & 597 & 51.7 \\
        NAS-MO-165 & 162.6 & 78.8 & 94.4 & 10.5 & 583 & 0.7 \\
        \sexyname-165 (Ours) & 162.7 & \textbf{80.3} & \textbf{95.0} & 10.5 & 582 & 0.7 \\
        \midrule
        NASNet-A~\cite{zoph2018learning} & 162.3 & 74.0 & 91.6 & 5.3 & 564 & $\sim$3 \\
        PONAS~\cite{huang2020ponas} & 182.4 & 75.2 & - & 5.6 & 376 & 8.8 \\
        EfficientNet B1~\cite{EfficientNet} & 192.7 & 79.2 & 94.5 & 7.8 & 700 & - \\
        OFA$^*$~\cite{cai2019once} & 204.3 & 80.2 & 95.1 & 9.1 & 743 & 51.7 \\
        {EVO}-190 & 188.1 & 79.5 & 94.8 & 11.3 & 687 & 0.7 \\
        NAS-MO-190 & 183.2 & 78.8 & 94.5 & 10.7 & 652 & 0.7 \\
        \sexyname-190 (Ours) & 185.5 & \textbf{80.4} & \textbf{95.0} & 10.4 & 640 & 0.7 \\
        \bottomrule[0.15em]
        \end{tabular}
    }
    \label{tab:gpu_comp}     
\end{table*}

\subsection{Architecture Search for CPU Devices}\label{sec:result_cpu}

We further exploit our PNAG to generate architectures under the latency budgets evaluated on a CPU device (Core i5-7400). Similar to the experiments for mobile devices, we evaluate our PNAG under 5 latency budgets, \ie, $\{$30ms, 35ms, 40ms, 45ms, 50ms$\}$.


As shown in Fig.~\ref{fig:cpu_compare}, our PNAG yields a large performance improvement over the considered two variants, \ie, EVO and NAS-MO, under diverse budgets. Moreover, our PNAG also outperforms popular NAS based (MnasNet, OFA$^*$) and manually designed architectures (MobileNetV2, MobileNetV3, and EfficientNet).
As for the quantitative comparisons, in Table~\ref{tab:cpu_comp}, our PNAG consistently yields the best results across all the considered latency budgets. To be specific, given a small latency budget $B{=}35$ms, our PNAG-35 yields better accuracy than the compared NAS methods with much lower search cost. Given a relatively large budget $B{=}50$ms, our PNAG-50 yields the same accuracy (80.5\%) as the best result on mobile devices (\ie, PNAG-200). This indicates that our PNAG generalizes well across the latency budgets based on different hardwares. 
Overall, these results demonstrate that our PNAG is able to generate very competitive architectures while satisfying diverse latency budgets.


\begin{table*}[t]
  \centering
  \caption{Comparisons of different reward functions based on \sexyname. We report the latency on mobile devices.
  }
    \resizebox{1.0\textwidth}{!}
    {      
    \begin{tabular}{c|cc|cc|cc|cc|cc}
    \toprule
    \multirow{2}[0]{*}{Reward}  & \multicolumn{2}{c|}{$B_1 {=} 80$ms} & \multicolumn{2}{c|}{$B_2 {=} 110$ms} & \multicolumn{2}{c|}{$B_3 {=} 140$ms} & \multicolumn{2}{c|}{$B_4 {=} 170$ms} & \multicolumn{2}{c}{$B_5 {=} 200$ms} \\
      & \multicolumn{1}{c}{Acc. (\%)} & \multicolumn{1}{c|}{Lat. (ms)} & \multicolumn{1}{c}{Acc.} & \multicolumn{1}{c|}{Lat. (ms)}  & \multicolumn{1}{c}{Acc.} & \multicolumn{1}{c|}{Lat. (ms)} & \multicolumn{1}{l}{Acc.} & \multicolumn{1}{c|}{Lat. (ms)} & \multicolumn{1}{l}{Acc.} & \multicolumn{1}{l}{Lat. (ms)} \\
    \hline
    \multirow{1}[0]{*}{Multi-objective Reward~\cite{tan2019mnasnet}} & 77.0   &  77.6  &   78.5  &  106.3    &   78.9  & 139.0  &   79.3   &  165.1  &  79.5    &   187.3  \\
    \multirow{1}[0]{*}{Multi-objective Absolute Reward~\cite{Bender2020TuNAS}} & 78.1   &  76.8   &   78.9  &  109.2 &   79.2   &  130.1    &   79.5   & 163.6  &  79.9  & 197.5 \\
    \multirow{1}[0]{*}{Pareto Dominance Reward (w/o acc. constraint)} &   73.8      &  74.4    &  73.6   &    64.9     &  74.3  &  66.5 & 73.9  &  70.0  & 74.0  &  70.8  \\
    \multirow{1}[0]{*}{Pareto Dominance Reward (Ours)} &    \textbf{78.4}  & 79.9    &  \textbf{79.5}  & 106.8  &   \textbf{79.8}  & 127.8 &  \textbf{80.3}  & 167.1    &   \textbf{80.5} & 193.9  \\
    
    \bottomrule
    \end{tabular}%
    }

  \label{tab:diff_reward}%
\end{table*}%

\begin{table*}[h]
  \centering
  \caption{Effect of different search strategies on the performance of \sexyname. We report the accuracy on ImageNet. 
  }
    \resizebox{0.7\textwidth}{!}
    {      
    \begin{tabular}{c|c|c|c|c|c}
    \toprule
    Search Strategy  & \multicolumn{1}{c|}{$B_1 {=} 80$ms} & \multicolumn{1}{c|}{$B_2 {=} 110$ms} & \multicolumn{1}{c|}{$B_3 {=} 140$ms} & \multicolumn{1}{c|}{$B_4 {=} 170$ms} & \multicolumn{1}{c}{$B_5 {=} 200$ms} \\
    \hline
    Repeated Independent Search  & 76.7 &   78.6       &   79.1     &   79.4       &  79.7         \\
    Pareto Frontier Search &    \textbf{78.4}      &  \textbf{79.5}    &   \textbf{79.8}  &   \textbf{80.3}      &   \textbf{80.5}   \\
    
    \bottomrule
    \end{tabular}%
    }
  \label{tab:pareto_learning}%
\end{table*}%

\begin{table}[t]
    \centering
    \caption{
    Comparisons of the time cost for architecture generation/design among different methods.}
    \resizebox{0.40\textwidth}{!}
    {      
    \begin{tabular}{c|cccc} 
    \toprule
    Method                &  \sexyname  & PC-DARTS   & ENAS  & DARTS  \\ 
    \midrule
    Time Cost &   \textbf{$\leq$5 s}   & 2 hours     & 12 hours & 4 days  \\
    \bottomrule
    \end{tabular}
    }

    \label{tab:generation_cost}
\end{table}

\subsection{Architecture Search for GPU Devices}\label{sec:result_gpu}

Besides the mobile and CPU devices, we also consider GPUs and adopt the latency on them as the computational budget. Since the inference speed on GPU is much faster than mobile processor and CPU, we measure the latency of deep models on a NVIDIA TITAN X GPU with a batch size of 64.
In this experiments, we compare different architecture design/search methods under the budgets of $\{$90ms, 115ms, 140ms, 165ms, 190ms$\}$.

As shown in Fig.~\ref{fig:gpu_compare}, similar to the results on mobile and CPU devices, our PNAG outperforms existing methods and the constructed variants by a large margin. We also reported the detailed comparisons in terms of accuracy and computational cost in Table~\ref{tab:gpu_comp}. Again, compared with both the hand-crafted methods (\eg, MobileNetV2~\cite{sandler2018mobilenetv2} and EfficientNet~\cite{EfficientNet}) and NAS methods (\eg, ENAS~\cite{pham2018efficient} and MnasNet~\cite{tan2019mnasnet}), our PNAG consistently produces better architectures under diverse budgets.
These results further emphasize the generalization ability of our PNAG to the latency budgets evaluated on different hardware devices.

\section{Further Experiments}
{In this section, we conduct ablation studies on our method.
Then we compare the architecture generation cost of our proposed method among different methods and discuss the impact of the number of considered budgets $K$.}

\subsection{Effect of the Pareto Dominance Reward}
We investigate the effectiveness of the Pareto frontier learning strategy and the Pareto dominance reward.
From Table~\ref{tab:diff_reward} and Table~\ref{tab:pareto_learning},
the Pareto frontier learning strategy tends to find better {architectures} than the independent search process due to the shared knowledge across the search processes under different budgets. 
Compared with two existing multi-objective rewards~\cite{tan2019mnasnet,Bender2020TuNAS},
the Pareto dominance reward encourages the generator to produce architectures that satisfy the considered budget constraints.
Moreover, if we do not consider accuracy constraint in the Pareto dominance reward, the generated architectures have low latency and poor accuracy.
With both the Pareto frontier learning strategy and the Pareto dominance reward, our method yields the best results under all budgets.  

\begin{table}[t]
  \centering
  \caption{Effect of $K$ on the generation performance of \sexyname. 
  We compare the {generated} architectures using different values of $K$ with the target latency $B{=}140$ms on ImageNet.}
  \resizebox{0.42\textwidth}{!}
   {
    \begin{tabular}{c|ccccc}
    \toprule
    $K$   & 1   &  2     & 5     & 10    & 30  \\
    \midrule
    Top-1 Acc. (\%) & 78.5 & 79.1 & 79.4 & \textbf{79.8} & \textbf{79.8}  \\
    \bottomrule
    \end{tabular}%
    }

  \label{tab:effect_k}%
\end{table}%

\subsection{Comparisons of Architecture Generation Cost}
{
In this part, we compare the architecture generation cost of different methods for 5 different budgets and show the comparison results in Table~\ref{tab:generation_cost}.
Given an arbitrary target budget, existing NAS methods need to perform an independent search to find feasible architectures. 
By contrast, since \sexyname directly learns the whole Pareto frontier, we are able to generate promising architectures based on a learned generator model via \emph{inference}.
Thus, the architecture generation cost of \sexyname is much less than other existing methods (See results in Table~\ref{tab:generation_cost}). In this sense, we are able to greatly accelerate the architecture design process in real-world scenarios.
These results demonstrate the efficiency of our \sexyname in generating architectures.
}

\subsection{Effect of $K$ on {the Generation Performance}}\label{sec:effect_k}
We investigate the effect of $K$ on the generation performance of \sexyname based on mobile device.
Note that we evenly select $K$ budgets from the range of latency. 
To investigate the effect of $K$, we consider several candidate values of {$K \in \{ 2, 5, 10, 30 \}$.}
We show the Top-1 accuracies of the architectures generated by \sexyname with different $K$ on ImageNet in Table~\ref{tab:effect_k}.
Since a small number of selected budgets $K$ cannot accurately approximate the ground-truth Pareto frontier or provide enough shared knowledge between different search processes, our method yields poor results with $K=2$.
When we increase $K$ larger than 5, we are able to greatly improve the performance of the generated architectures. From Table~\ref{tab:effect_k}, our method yields the best result when $K\geq10$ and we use this setting in the experiments.

\section{Conclusion}
In this paper, 
we focus on designing effective and feasible architectures via an architecture generation process. To this end, we have proposed a novel Pareto-aware Neural Architecture Generator (\sexyname) which only needs to be trained once and dynamically generates promising architectures satisfying any given budget via inference.
Unlike existing methods, we seek to learn the whole Pareto frontier instead of finding a single or several discrete Pareto optimal architectures.
Based on the learned Pareto frontier, our \sexyname consistently outperforms existing NAS methods across diverse budgets. 
Extensive experiments on three hardware platforms (\ie, mobile devices, CPU, and GPU) demonstrate the effectiveness of the proposed method.

\bibliographystyle{IEEEtran}
\bibliography{mybib}

\begin{thebibliography}{10}
\providecommand{\url}[1]{#1}
\csname url@samestyle\endcsname
\providecommand{\newblock}{\relax}
\providecommand{\bibinfo}[2]{#2}
\providecommand{\BIBentrySTDinterwordspacing}{\spaceskip=0pt\relax}
\providecommand{\BIBentryALTinterwordstretchfactor}{4}
\providecommand{\BIBentryALTinterwordspacing}{\spaceskip=\fontdimen2\font plus
\BIBentryALTinterwordstretchfactor\fontdimen3\font minus
  \fontdimen4\font\relax}
\providecommand{\BIBforeignlanguage}[2]{{%
\expandafter\ifx\csname l@#1\endcsname\relax
\typeout{** WARNING: IEEEtran.bst: No hyphenation pattern has been}%
\typeout{** loaded for the language `#1'. Using the pattern for}%
\typeout{** the default language instead.}%
\else
\language=\csname l@#1\endcsname
\fi
#2}}
\providecommand{\BIBdecl}{\relax}
\BIBdecl

\bibitem{lecun1989backpropagation}
Y.~LeCun, B.~Boser, J.~S. Denker, D.~Henderson, R.~E. Howard, W.~Hubbard, and
  L.~D. Jackel, ``{B}ackpropagation applied to handwritten zip code
  recognition,'' \emph{Neural Computation}, vol.~1, no.~4, pp. 541--551, 1989.

\bibitem{krizhevsky2012imagenet}
A.~Krizhevsky, I.~Sutskever, and G.~E. Hinton, ``Imagenet classification with
  deep convolutional neural networks,'' in \emph{Advances in Neural Information
  Processing Systems}, 2012, pp. 1097--1105.

\bibitem{srivastava2015training}
R.~K. Srivastava, K.~Greff, and J.~Schmidhuber, ``{T}raining very deep
  networks,'' in \emph{Advances in Neural Information Processing Systems},
  2015, pp. 2377--2385.

\bibitem{he2016deep}
K.~He, X.~Zhang, S.~Ren, and J.~Sun, ``Deep residual learning for image
  recognition,'' in \emph{IEEE Conference on Computer Vision and Pattern
  Recognition}, 2016, pp. 770--778.

\bibitem{alexey2021vit}
A.~Dosovitskiy, L.~Beyer, A.~Kolesnikov, D.~Weissenborn, X.~Zhai,
  T.~Unterthiner, M.~Dehghani, M.~Minderer, G.~Heigold, S.~Gelly, J.~Uszkoreit,
  and N.~Houlsby, ``An image is worth 16x16 words: Transformers for image
  recognition at scale,'' in \emph{International Conference on Learning
  Representations}, 2021.

\bibitem{liu2021swin}
Z.~Liu, Y.~Lin, Y.~Cao, H.~Hu, Y.~Wei, Z.~Zhang, S.~Lin, and B.~Guo, ``Swin
  transformer: Hierarchical vision transformer using shifted windows,'' in
  \emph{IEEE International Conference on Computer Vision}, 2021, pp.
  9992--10\,002.

\bibitem{evan2017fully}
E.~Shelhamer, J.~Long, and T.~Darrell, ``Fully convolutional networks for
  semantic segmentation,'' \emph{IEEE Transactions on Pattern Analysis and
  Machine Intelligence}, vol.~39, no.~4, pp. 640--651, 2017.

\bibitem{chen2018encoder}
L.~Chen, Y.~Zhu, G.~Papandreou, F.~Schroff, and H.~Adam, ``Encoder-decoder with
  atrous separable convolution for semantic image segmentation,'' in
  \emph{European Conference on Computer Vision}, V.~Ferrari, M.~Hebert,
  C.~Sminchisescu, and Y.~Weiss, Eds., vol. 11211, 2018, pp. 833--851.

\bibitem{xie2021segformer}
E.~Xie, W.~Wang, Z.~Yu, A.~Anandkumar, J.~M. Alvarez, and P.~Luo, ``Segformer:
  Simple and efficient design for semantic segmentation with transformers,''
  \emph{Advances in Neural Information Processing Systems}, vol.~34, 2021.

\bibitem{wang2021hrnet}
J.~Wang, K.~Sun, T.~Cheng, B.~Jiang, C.~Deng, Y.~Zhao, D.~Liu, Y.~Mu, M.~Tan,
  X.~Wang, W.~Liu, and B.~Xiao, ``Deep high-resolution representation learning
  for visual recognition,'' \emph{IEEE Transactions on Pattern Analysis and
  Machine Intelligence}, vol.~43, no.~10, pp. 3349--3364, 2021.

\bibitem{joseph2018yolov3}
J.~Redmon and A.~Farhadi, ``Yolov3: An incremental improvement,'' \emph{arXiv
  preprint arXiv:1804.02767}, 2018.

\bibitem{zhao2019object}
Z.-Q. Zhao, P.~Zheng, S.-t. Xu, and X.~Wu, ``Object detection with deep
  learning: A review,'' \emph{IEEE Transactions on Neural Networks and Learning
  Systems}, vol.~30, no.~11, pp. 3212--3232, 2019.

\bibitem{tian2019fcos}
Z.~Tian, C.~Shen, H.~Chen, and T.~He, ``Fcos: Fully convolutional one-stage
  object detection,'' in \emph{IEEE International Conference on Computer
  Vision}, 2019, pp. 9627--9636.

\bibitem{nicolas2020end}
N.~Carion, F.~Massa, G.~Synnaeve, N.~Usunier, A.~Kirillov, and S.~Zagoruyko,
  ``End-to-end object detection with transformers,'' in \emph{European
  Conference on Computer Vision}, A.~Vedaldi, H.~Bischof, T.~Brox, and
  J.~Frahm, Eds., vol. 12346, 2020, pp. 213--229.

\bibitem{zoph2016neural}
B.~Zoph and Q.~V. Le, ``Neural architecture search with reinforcement
  learning,'' in \emph{International Conference on Learning Representations},
  2017.

\bibitem{zoph2018learning}
B.~Zoph, V.~Vasudevan, J.~Shlens, and Q.~V. Le, ``Learning transferable
  architectures for scalable image recognition,'' in \emph{IEEE Conference on
  Computer Vision and Pattern Recognition}, 2018, pp. 8697--8710.

\bibitem{li2020block}
C.~Li, J.~Peng, L.~Yuan, G.~Wang, X.~Liang, L.~Lin, and X.~Chang,
  ``Block-wisely supervised neural architecture search with knowledge
  distillation,'' in \emph{IEEE Conference on Computer Vision and Pattern
  Recognition}, 2020, pp. 1989--1998.

\bibitem{tan2020efficientdet}
M.~Tan, R.~Pang, and Q.~V. Le, ``Efficientdet: Scalable and efficient object
  detection,'' in \emph{IEEE Conference on Computer Vision and Pattern
  Recognition}, 2020, pp. 10\,778--10\,787.

\bibitem{dai2021fbnetv3}
X.~Dai, A.~Wan, P.~Zhang, B.~Wu, Z.~He, Z.~Wei, K.~Chen, Y.~Tian, M.~Yu,
  P.~Vajda, and J.~E. Gonzalez, ``Fbnetv3: Joint architecture-recipe search
  using predictor pretraining,'' in \emph{IEEE Conference on Computer Vision
  and Pattern Recognition}, June 2021, pp. 16\,276--16\,285.

\bibitem{white2021powerful}
C.~White, A.~Zela, R.~Ru, Y.~Liu, and F.~Hutter, ``How powerful are performance
  predictors in neural architecture search?'' \emph{Advances in Neural
  Information Processing Systems}, vol.~34, 2021.

\bibitem{chen2021neural}
W.~Chen, X.~Gong, and Z.~Wang, ``Neural architecture search on imagenet in four
  gpu hours: A theoretically inspired perspective,'' \emph{arXiv preprint
  arXiv:2102.11535}, 2021.

\bibitem{yan2021fp}
Z.~Yan, X.~Dai, P.~Zhang, Y.~Tian, B.~Wu, and M.~Feiszli, ``Fp-nas: Fast
  probabilistic neural architecture search,'' in \emph{Proceedings of the
  IEEE/CVF Conference on Computer Vision and Pattern Recognition}, 2021, pp.
  15\,139--15\,148.

\bibitem{guo2020breaking}
Y.~Guo, Y.~Chen, Y.~Zheng, P.~Zhao, J.~Chen, J.~Huang, and M.~Tan, ``Breaking
  the curse of space explosion: Towards efficient nas with curriculum search,''
  in \emph{International Conference on Machine Learning}, 2020.

\bibitem{tan2019mnasnet}
M.~Tan, B.~Chen, R.~Pang, V.~Vasudevan, M.~Sandler, A.~Howard, and Q.~V. Le,
  ``Mnasnet: Platform-aware neural architecture search for mobile,'' in
  \emph{IEEE Conference on Computer Vision and Pattern Recognition}, 2019, pp.
  2820--2828.

\bibitem{stamoulis2019single}
D.~Stamoulis, R.~Ding, D.~Wang, D.~Lymberopoulos, B.~Priyantha, J.~Liu, and
  D.~Marculescu, ``Single-path nas: Designing hardware-efficient convnets in
  less than 4 hours,'' in \emph{Joint European Conference on Machine Learning
  and Knowledge Discovery in Databases}, 2019, pp. 481--497.

\bibitem{lu2019nsga}
Z.~Lu, I.~Whalen, V.~Boddeti, Y.~Dhebar, K.~Deb, E.~Goodman, and W.~Banzhaf,
  ``Nsga-net: neural architecture search using multi-objective genetic
  algorithm,'' in \emph{Proceedings of the Genetic and Evolutionary Computation
  Conference}, 2019, pp. 419--427.

\bibitem{lu2020nsganetv2}
Z.~Lu, K.~Deb, E.~Goodman, W.~Banzhaf, and V.~N. Boddeti, ``Nsganetv2:
  Evolutionary multi-objective surrogate-assisted neural architecture search,''
  in \emph{European Conference on Computer Vision}.\hskip 1em plus 0.5em minus
  0.4em\relax Springer, 2020, pp. 35--51.

\bibitem{kim2005adaptive}
I.~Y. Kim and O.~L. De~Weck, ``Adaptive weighted-sum method for bi-objective
  optimization: Pareto front generation,'' \emph{Structural and
  Multidisciplinary Optimization}, vol.~29, no.~2, pp. 149--158, 2005.

\bibitem{he2020milenas}
C.~He, H.~Ye, L.~Shen, and T.~Zhang, ``Milenas: Efficient neural architecture
  search via mixed-level reformulation,'' in \emph{IEEE Conference on Computer
  Vision and Pattern Recognition}, 2020, pp. 11\,990--11\,999.

\bibitem{li2020sgas}
G.~Li, G.~Qian, I.~C. Delgadillo, M.~M{\"{u}}ller, A.~K. Thabet, and B.~Ghanem,
  ``{SGAS:} sequential greedy architecture search,'' in \emph{IEEE Conference
  on Computer Vision and Pattern Recognition}, 2020, pp. 1617--1627.

\bibitem{yang2021netadaptv2}
T.-J. Yang, Y.-L. Liao, and V.~Sze, ``Netadaptv2: Efficient neural architecture
  search with fast super-network training and architecture optimization,'' in
  \emph{IEEE Conference on Computer Vision and Pattern Recognition}, 2021, pp.
  2402--2411.

\bibitem{zhang2021you}
X.~Zhang, Z.~Huang, N.~Wang, S.~Xiang, and C.~Pan, ``You only search once:
  Single shot neural architecture search via direct sparse optimization,''
  \emph{IEEE Transactions on Pattern Analysis and Machine Intelligence},
  vol.~43, no.~9, pp. 2891--2904, 2021.

\bibitem{zheng2021migonas}
X.~Zheng, R.~Ji, Y.~Chen, Q.~Wang, B.~Zhang, J.~Chen, Q.~Ye, F.~Huang, and
  Y.~Tian, ``{MIGO-NAS:} towards fast and generalizable neural architecture
  search,'' \emph{IEEE Transactions on Pattern Analysis and Machine
  Intelligence}, vol.~43, no.~9, pp. 2936--2952, 2021.

\bibitem{pham2018efficient}
H.~Pham, M.~Guan, B.~Zoph, Q.~Le, and J.~Dean, ``Efficient neural architecture
  search via parameter sharing,'' in \emph{International Conference on Machine
  Learning}, 2018, pp. 4095--4104.

\bibitem{pasunuru2019continual}
R.~Pasunuru and M.~Bansal, ``Continual and multi-task architecture search,'' in
  \emph{Proceedings of Conference of the Association for Computational
  Linguistics}, A.~Korhonen, D.~R. Traum, and L.~M{\`{a}}rquez, Eds., 2019, pp.
  1911--1922.

\bibitem{tian2020offrl}
Y.~Tian, Q.~Wang, Z.~Huang, W.~Li, D.~Dai, M.~Yang, J.~Wang, and O.~Fink,
  ``Off-policy reinforcement learning for efficient and effective {GAN}
  architecture search,'' in \emph{European Conference on Computer Vision},
  A.~Vedaldi, H.~Bischof, T.~Brox, and J.~Frahm, Eds., vol. 12352, 2020, pp.
  175--192.

\bibitem{arash2020unas}
A.~Vahdat, A.~Mallya, M.~Liu, and J.~Kautz, ``{UNAS:} differentiable
  architecture search meets reinforcement learning,'' in \emph{IEEE Conference
  on Computer Vision and Pattern Recognition}, 2020, pp. 11\,263--11\,272.

\bibitem{real2017large}
E.~Real, S.~Moore, A.~Selle, S.~Saxena, Y.~L. Suematsu, J.~Tan, Q.~V. Le, and
  A.~Kurakin, ``Large-scale evolution of image classifiers,'' in
  \emph{International Conference on Machine Learning}, 2017, pp. 2902--2911.

\bibitem{real2019regularized}
E.~Real, A.~Aggarwal, Y.~Huang, and Q.~V. Le, ``Regularized evolution for image
  classifier architecture search,'' in \emph{AAAI Conference on Artificial
  Intelligence}, vol.~33, 2019, pp. 4780--4789.

\bibitem{lu2021neural}
Z.~Lu, G.~Sreekumar, E.~Goodman, W.~Banzhaf, K.~Deb, and V.~N. Boddeti,
  ``Neural architecture transfer,'' \emph{IEEE Transactions on Pattern Analysis
  and Machine Intelligence}, vol.~43, no.~9, pp. 2971--2989, 2021.

\bibitem{ming_zennas_iccv2021}
M.~Lin, P.~Wang, Z.~Sun, H.~Chen, X.~Sun, Q.~Qian, H.~Li, and R.~Jin,
  ``Zen-nas: A zero-shot nas for high-performance deep image recognition,'' in
  \emph{IEEE International Conference on Computer Vision}, 2021.

\bibitem{chen2021autoformer}
M.~Chen, H.~Peng, J.~Fu, and H.~Ling, ``Autoformer: Searching transformers for
  visual recognition,'' in \emph{IEEE International Conference on Computer
  Vision}, 2021, pp. 12\,250--12\,260.

\bibitem{liu2021survey}
Y.~Liu, Y.~Sun, B.~Xue, M.~Zhang, G.~G. Yen, and K.~C. Tan, ``A survey on
  evolutionary neural architecture search,'' \emph{IEEE transactions on neural
  networks and learning systems}, 2021.

\bibitem{liu2018darts}
H.~Liu, K.~Simonyan, and Y.~Yang, ``Darts: Differentiable architecture
  search,'' in \emph{International Conference on Learning Representations},
  2019.

\bibitem{chen2019progressive}
X.~Chen, L.~Xie, J.~Wu, and Q.~Tian, ``Progressive differentiable architecture
  search: Bridging the depth gap between search and evaluation,'' in \emph{IEEE
  International Conference on Computer Vision}, 2019, pp. 1294--1303.

\bibitem{xu2020pcdarts}
Y.~Xu, L.~Xie, X.~Zhang, X.~Chen, G.-J. Qi, Q.~Tian, and H.~Xiong, ``Pc-darts:
  Partial channel connections for memory-efficient differentiable architecture
  search,'' in \emph{International Conference on Learning Representations},
  2020.

\bibitem{chu2021dartsminus}
X.~Chu, X.~Wang, B.~Zhang, S.~Lu, X.~Wei, and J.~Yan, ``{DARTS-:} robustly
  stepping out of performance collapse without indicators,'' in
  \emph{International Conference on Learning Representations}, 2021.

\bibitem{chen2021drnas}
X.~Chen, R.~Wang, M.~Cheng, X.~Tang, and C.~Hsieh, ``Drnas: Dirichlet neural
  architecture search,'' in \emph{International Conference on Learning
  Representations}, 2021.

\bibitem{guo2022towards}
Y.~Guo, J.~Wang, Q.~Chen, J.~Cao, Z.~Deng, Y.~Xu, J.~Chen, and M.~Tan,
  ``Towards lightweight super-resolution with dual regression learning,''
  \emph{arXiv preprint arXiv:2207.07929}, 2022.

\bibitem{zhao2021few}
Y.~Zhao, L.~Wang, Y.~Tian, R.~Fonseca, and T.~Guo, ``Few-shot neural
  architecture search,'' in \emph{International Conference on Machine
  Learning}.\hskip 1em plus 0.5em minus 0.4em\relax PMLR, 2021, pp.
  12\,707--12\,718.

\bibitem{yi2021renas}
Y.~Xu, Y.~Wang, K.~Han, Y.~Tang, S.~Jui, C.~Xu, and C.~Xu, ``Renas:
  Relativistic evaluation of neural architecture search,'' in \emph{IEEE
  Conference on Computer Vision and Pattern Recognition}, 2021, pp. 4411--4420.

\bibitem{chu2021fairnas}
X.~Chu, B.~Zhang, and R.~Xu, ``Fairnas: Rethinking evaluation fairness of
  weight sharing neural architecture search,'' in \emph{IEEE International
  Conference on Computer Vision}.\hskip 1em plus 0.5em minus 0.4em\relax
  {IEEE}, 2021, pp. 12\,219--12\,228.

\bibitem{xie2019exploring}
S.~Xie, A.~Kirillov, R.~Girshick, and K.~He, ``Exploring randomly wired neural
  networks for image recognition,'' in \emph{IEEE International Conference on
  Computer Vision}, 2019, pp. 1284--1293.

\bibitem{ru2020neural}
B.~Ru, P.~Esperanca, and F.~Carlucci, ``Neural architecture generator
  optimization,'' in \emph{Advances in Neural Information Processing Systems},
  2020.

\bibitem{cai2019once}
H.~Cai, C.~Gan, and S.~Han, ``Once for all: Train one network and specialize it
  for efficient deployment,'' in \emph{International Conference on Learning
  Representations}, 2020.

\bibitem{huang2020ponas}
S.-Y. Huang and W.-T. Chu, ``Ponas: Progressive one-shot neural architecture
  search for very efficient deployment,'' \emph{arXiv preprint
  arXiv:2003.05112}, 2020.

\bibitem{elsken2018efficient}
J.~H.~M. Thomas~Elsken and F.~Hutter, ``Efficient multi-objective neural
  architecture search via lamarckian evolution,'' in \emph{International
  Conference on Learning Representations}, 2019.

\bibitem{Bender2020TuNAS}
G.~Bender, H.~Liu, B.~Chen, G.~Chu, S.~Cheng, P.~Kindermans, and Q.~V. Le,
  ``Can weight sharing outperform random architecture search? an investigation
  with tunas,'' in \emph{IEEE International Conference on Computer Vision},
  2020, pp. 14\,311--14\,320.

\bibitem{guo2020single}
Z.~Guo, X.~Zhang, H.~Mu, W.~Heng, Z.~Liu, Y.~Wei, and J.~Sun, ``Single path
  one-shot neural architecture search with uniform sampling,'' in
  \emph{European Conference on Computer Vision}, 2020, pp. 544--560.

\bibitem{li2021hw}
C.~Li, Z.~Yu, Y.~Fu, Y.~Zhang, Y.~Zhao, H.~You, Q.~Yu, Y.~Wang, and Y.~Lin,
  ``Hw-nas-bench: Hardware-aware neural architecture search benchmark,''
  \emph{arXiv preprint arXiv:2103.10584}, 2021.

\bibitem{huang2021searching}
S.-Y. Huang and W.-T. Chu, ``Searching by generating: Flexible and efficient
  one-shot nas with architecture generator,'' in \emph{IEEE Conference on
  Computer Vision and Pattern Recognition}, 2021, pp. 983--992.

\bibitem{deb2002fast}
K.~Deb, A.~Pratap, S.~Agarwal, and T.~Meyarivan, ``A fast and elitist
  multiobjective genetic algorithm: Nsga-ii,'' \emph{IEEE Transactions on
  Evolutionary Computation}, vol.~6, no.~2, pp. 182--197, 2002.

\bibitem{kim2004spea}
M.~Kim, T.~Hiroyasu, M.~Miki, and S.~Watanabe, ``{SPEA2+:} improving the
  performance of the strength pareto evolutionary algorithm 2,'' in
  \emph{Parallel Problem Solving from Nature}, vol. 3242.\hskip 1em plus 0.5em
  minus 0.4em\relax Springer, 2004, pp. 742--751.

\bibitem{cheng2018searching}
A.-C. Cheng, J.-D. Dong, C.-H. Hsu, S.-H. Chang, M.~Sun, S.-C. Chang, J.-Y.
  Pan, Y.-T. Chen, W.~Wei, and D.-C. Juan, ``Searching toward pareto-optimal
  device-aware neural architectures,'' in \emph{Proceedings of the
  International Conference on Computer-Aided Design}, 2018, pp. 1--7.

\bibitem{dong2018dpp}
J.-D. Dong, A.-C. Cheng, D.-C. Juan, W.~Wei, and M.~Sun, ``Dpp-net:
  Device-aware progressive search for pareto-optimal neural architectures,'' in
  \emph{European Conference on Computer Vision}, 2018, pp. 517--531.

\bibitem{lu2020multi}
Z.~Lu, I.~Whalen, Y.~Dhebar, K.~Deb, E.~Goodman, W.~Banzhaf, and V.~N. Boddeti,
  ``Multi-objective evolutionary design of deep convolutional neural networks
  for image classification,'' \emph{IEEE Transactions on Evolutionary
  Computation}, 2020.

\bibitem{grosan2008generating}
C.~Grosan and A.~Abraham, ``Generating uniformly distributed pareto optimal
  points for constrained and unconstrained multicriteria optimization,''
  \emph{International Conference on Informatics and Systems}, pp. 27--29, 2008.

\bibitem{guo2019nat}
Y.~Guo, Y.~Zheng, M.~Tan, Q.~Chen, J.~Chen, P.~Zhao, and J.~Huang, ``{NAT}:
  Neural architecture transformer for accurate and compact architectures,'' in
  \emph{Advances in Neural Information Processing Systems}, 2019.

\bibitem{guo2021towards}
Y.~Guo, Y.~Zheng, M.~Tan, Q.~Chen, Z.~Li, J.~Chen, P.~Zhao, and J.~Huang,
  ``Towards accurate and compact architectures via neural architecture
  transformer,'' \emph{IEEE Transactions on Pattern Analysis and Machine
  Intelligence}, 2021.

\bibitem{freund2003efficient}
Y.~Freund, R.~Iyer, R.~E. Schapire, and Y.~Singer, ``An efficient boosting
  algorithm for combining preferences,'' \emph{Journal of Machine Learning
  Research}, vol.~4, no. Nov, pp. 933--969, 2003.

\bibitem{burges2005learning}
C.~Burges, T.~Shaked, E.~Renshaw, A.~Lazier, M.~Deeds, N.~Hamilton, and
  G.~Hullender, ``Learning to rank using gradient descent,'' in
  \emph{International Conference on Machine Learning}, 2005, pp. 89--96.

\bibitem{chen2009ranking}
W.~Chen, T.-Y. Liu, Y.~Lan, Z.-M. Ma, and H.~Li, ``Ranking measures and loss
  functions in learning to rank,'' in \emph{Advances in Neural Information
  Processing Systems}, 2009, pp. 315--323.

\bibitem{howard2019searching}
A.~Howard, M.~Sandler, G.~Chu, L.-C. Chen, B.~Chen, M.~Tan, W.~Wang, Y.~Zhu,
  R.~Pang, V.~Vasudevan \emph{et~al.}, ``Searching for mobilenetv3,'' in
  \emph{IEEE International Conference on Computer Vision}, 2019, pp.
  1314--1324.

\bibitem{sandler2018mobilenetv2}
M.~Sandler, A.~Howard, M.~Zhu, A.~Zhmoginov, and L.-C. Chen, ``Mobilenetv2:
  Inverted residuals and linear bottlenecks,'' in \emph{IEEE Conference on
  Computer Vision and Pattern Recognition}, 2018, pp. 4510--4520.

\bibitem{wan2020fbnetv2}
A.~Wan, X.~Dai, P.~Zhang, Z.~He, Y.~Tian, S.~Xie, B.~Wu, M.~Yu, T.~Xu, K.~Chen
  \emph{et~al.}, ``Fbnetv2: Differentiable neural architecture search for
  spatial and channel dimensions,'' in \emph{IEEE Conference on Computer Vision
  and Pattern Recognition}, 2020, pp. 12\,965--12\,974.

\bibitem{wu2019fbnet}
B.~Wu, X.~Dai, P.~Zhang, Y.~Wang, F.~Sun, Y.~Wu, Y.~Tian, P.~Vajda, Y.~Jia, and
  K.~Keutzer, ``Fbnet: Hardware-aware efficient convnet design via
  differentiable neural architecture search,'' in \emph{IEEE Conference on
  Computer Vision and Pattern Recognition}, 2019, pp. 10\,734--10\,742.

\bibitem{cai2018proxylessnas}
H.~Cai, L.~Zhu, and S.~Han, ``Proxyless{NAS}: Direct neural architecture search
  on target task and hardware,'' in \emph{International Conference on Learning
  Representations}, 2019.

\bibitem{yu2020bignas}
J.~Yu, P.~Jin, H.~Liu, G.~Bender, P.-J. Kindermans, M.~Tan, T.~Huang, X.~Song,
  R.~Pang, and Q.~Le, ``Bignas: Scaling up neural architecture search with big
  single-stage models,'' in \emph{European Conference on Computer
  Vision}.\hskip 1em plus 0.5em minus 0.4em\relax Springer, 2020, pp. 702--717.

\bibitem{EfficientNet}
M.~Tan and Q.~V. Le, ``Efficientnet: Rethinking model scaling for convolutional
  neural networks,'' in \emph{International Conference on Machine Learning},
  2019, pp. 6105--6114.

\bibitem{peng2020cream}
H.~Peng, H.~Du, H.~Yu, Q.~Li, J.~Liao, and J.~Fu, ``Cream of the crop:
  Distilling prioritized paths for one-shot neural architecture search,'' in
  \emph{Advances in Neural Information Processing Systems}, 2020.

\end{thebibliography}

\appendices
	

\end{document}